\documentclass[10pt,twocolumn]{article} 
\usepackage{arxiv}

\usepackage[utf8]{inputenc} 
\usepackage[T1]{fontenc}    
\usepackage{hyperref}       
\usepackage{url}            
\usepackage{booktabs}       %
\usepackage{amsfonts}       
\usepackage{nicefrac}       
\usepackage{amsmath}
\usepackage{graphicx}
\usepackage{enumitem}
\usepackage{color,soul}
\usepackage{subcaption}
\usepackage[english]{babel}
\usepackage[dvipsnames]{xcolor}
\usepackage{geometry}
\usepackage{atbegshi}
\usepackage{graphicx}
\usepackage{textcomp}
\usepackage{hyperref}
\usepackage[titlenumbered,ruled]{algorithm2e}
\usepackage{microtype}     
\usepackage{graphicx}
\usepackage{natbib}

\title{Hybrid Classifiers for Spatio-temporal Real-time Abnormal Behaviors Detection, Tracking, and Recognition in Massive Hajj Crowds}

\author{Tarik~Alafif\\ Department of Computer Science\\ Jamoum University College\\ Umm Al-Qura University\\ Jamoum, Makkah, Saudi Arabia 25375\\
	\And
        Anas~Hadi\\ Faculty of Computing and Information Technology\\ King Abdulaziz University\\ Jeddah, Saudi Arabia\\
        \And
        Manal~Allahyani\\ Faculty of Computing and Information Technology\\ King Abdulaziz University\\ Jeddah, Saudi Arabia\\
        \And
        Bander~Alzahrani\\Faculty of Computing and Information Technology\\ King Abdulaziz University\\ Jeddah, Saudi Arabia\\
        \And
        Areej~Alhothali\\  Faculty of Computing and Information Technology\\ King Abdulaziz University\\ Jeddah, Saudi Arabia\\
         \And
        Reem~Alotaibi\\Faculty of Computing and Information Technology\\ King Abdulaziz University\\ Jeddah, Saudi Arabia\\
          \And
        Ahmed~Barnawi\\  Faculty of Computing and Information Technology\\ King Abdulaziz University\\ Jeddah, Saudi Arabia\\
}

\date{}

\renewcommand{\shorttitle}{Hybrid Classifiers for Abnormal Behaviors Detection in Massive Hajj Crowds}

\begin{document}
\twocolumn[
  \begin{@twocolumnfalse}
    \maketitle
      \begin{abstract}
Individual abnormal behaviors vary depending on crowd sizes, contexts, and scenes. Challenges such as partial occlusions, blurring, large-number abnormal behavior, and camera viewing occur in large-scale crowds when detecting, tracking, and recognizing individuals with abnormal behaviors. In this paper, our contribution is twofold. First, we introduce an annotated and labeled large-scale crowd abnormal behaviors Hajj dataset (HAJJv2). Second, we propose two methods of hybrid Convolutional Neural Networks (CNNs) and Random Forests (RFs) to detect and recognize Spatio-temporal abnormal behaviors in small and large-scales crowd videos. In small-scale crowd videos, a ResNet-50 pre-trained CNN model is fine-tuned to verify whether every frame is normal or abnormal in the spatial domain. If anomalous behaviors are observed, a motion-based individuals detection method based on the magnitudes and orientations of Horn-Schunck optical flow is used to locate and track individuals with abnormal behaviors. A Kalman filter is employed in large-scale crowd videos to predict and track the detected individuals in the subsequent frames. Then, means, variances, and standard deviations statistical features are computed and fed to the RF to classify individuals with abnormal behaviors in the temporal domain. In large-scale crowds, we fine-tune the ResNet-50 model using YOLOv2 object detection technique to detect individuals with abnormal behaviors in the spatial domain. The proposed method achieves 99.77\% and 93.71\% of average Area Under the Curves (AUCs) on two public benchmark small-scale crowd datasets, UMN and UCSD, respectively. While the large-scale crowd method achieves 76.08\% average AUC using the HAJJv2 dataset. Our method outperforms state-of-the-art methods using the small-scale crowd datasets with a margin of 1.67\%, 6.06\%, and 2.85\% on UMN, UCSD Ped1, and UCSD Ped2, respectively. It also achieves a satisfactory result in the large-scale crowds.
\end{abstract}

  \end{@twocolumnfalse}
]

\section{Introduction}
\label{sec:introduction}
Abnormal behavior detection in videos has been receiving lots of attention. This research area has been widely examined in the past two decades due to its importance and challenging nature in the computer vision domain. Generally, abnormal behavior is described as the unusual act of an individual in an event such as running, walking in the opposite direction, jumping, etc. Individual abnormal behaviors can be perceived differently according to crowd contexts and scenes. Therefore, the definition of abnormal behaviors might vary from one place or scenario to another. Similarly, the density and the number of individuals in the crowd often vary significantly, which can be small or large-scale crowds according to the context of the scene. A small-scale crowd often contains approximately tens of individuals gathering or moving in the same location, while a large-scale crowd contains hundreds or thousands of individuals in the same place. Therefore, the large-scale crowd scene might raise many challenges as a result of many individuals moving or gathering in one location at almost the same time. In large-scale crowds, challenges such as partial and full occlusions, blurring, large-number abnormal behaviors, and low scaling usually occur when detecting, tracking, and recognizing abnormal behaviors. As a result, detecting, tracking, and recognizing anomalous actions in large crowds are difficult, whereas performing comparable tasks in small crowds is easier.  

To ensure safety in public places, many studies have tackled the problem of abnormality detection in crowd scenes. These studies have exploited wide range of trajectory features~\citep{zhang2016exploring,Aniket2016,7178180,8551517,cocsar2016toward,piciarelli2008trajectory}, dense motion features~\citep{colque2016histograms,CHO201464,qasim2019hybrid,zhang2014social}, spatial-temporal features~\citep{10.1016/j.neucom.2015.07.153,6844850,fradi2016crowd}, or deep learning based features and optimization techniques for anomaly recognition~\citep{7234886,SIKDAR2020317,9559976,bansod2019anomalous,7858798}. Most of the developed methods perform a binary frame-level for anomaly detection. Several studies considered locating anomalies in crowd surveillance videos~\citep{bansod2019anomalous,CHAKER2017266,ZHOU2016358,6985609,7858798,bansod2020crowd}, and less attention was paid to multi-class anomalies~\citep{sikdar2019multi}. This study proposes a hybrid model that first identifies anomaly at the frame-level and then locates and classifies crowd anomaly into one of multiple classes.
 Distinguishing between different types of abnormal behaviors (e.g., running and walking against the crowd) raises many challenges that are worth research attention. The proposed methods in the field are also often evaluated on datasets of low to moderate crowd density levels. In this research, we evaluate the proposed methods on both moderate and very high crowd density levels of benchmark datasets and HAJJ dataset.

Hajj is an annual religious pilgrimage that takes place in Makkah, Saudi Arabia. It is considered a large-scale event because it regularly attracts over two million pilgrims from various countries and continents who congregate in one location. The diversity and cultural differences of pilgrims reduce the ability to understand their abnormal behaviors. However, we define, annotate, and label a set of abnormal behaviors based on the context of the Hajj. The definition of abnormal behaviors has been studied thoroughly in this research and is associated with the causes of potential obstacles or dangers to large-scale crowd flows. 
This analysis aims to help automate the detection, tracking, and recognition of abnormal behaviors for large-scale crowds in surveillance cameras to ensure pilgrims' safety and smooth flow during Hajj. It also helps security authorities and decision-makers to visualize and anticipate potential risks.

Our work is inspired by the power of Convolutional Neural Networks (CNNs) and transfer learning in many computer vision tasks~\citep{lecun1995convolutional, lecun1998gradient, he2016deep}. In addition to the success of CNNs, the work is also motivated by the success of Random Forests (RFs) in the classification of unstructured data~\citep{liaw2002classification}. The contributions of this work are summarized as follows:
\begin{itemize}
\item We introduce a manually annotated and labeled large-scale crowd abnormal behaviors dataset in Hajj, HAJJv2.

\item We propose two methods of hybrid CNNs and RFs classifiers to detect, track, and recognize Spatio-temporal abnormal behaviors in small-scale and large-scale crowd videos.

\item We evaluate the first proposed method on two common benchmark abnormal behaviors public small-scale crowd video datasets, UMN and UCSD, against the currently published methods. Then, we evaluate the second proposed method on the HAJJv2 dataset and compare it with the previously existing method.

\end{itemize}

The remainder of this paper is organized as follows. We provide a literature review for abnormal behaviors detection and recognition in {\bf Section \ref{sec:Related}}. In {\bf Section \ref{sec:Hajj}}, we briefly describe the abnormal behavior HAJJv2 dataset. Then, we present our proposed methods to detect, track, and recognize Spatio-temporal abnormal behaviors in small-scale and large-scale crowd videos in {\bf Section \ref{sec:meth}}. Experimental implementation, results, and evaluation are provided in {\bf Section \ref{sec:Exp}}. Then, a discussion on experimental evaluations, limitations, and challenges is provided in {\bf Section \ref{sec:Dis}}. Finally, we conclude our work in {\bf Section \ref{sec:Con}}.

\section{Related work}\label{sec:Related} Many research works have been proposed to detect abnormal behaviors in crowds in the past two decades. In this section, we provide the most recent related work. Current abnormal behaviors detection and recognition methods can be briefly overviewed in two scales of crowds as follows: 
\begin{itemize}
\item \textbf{Small-scale crowds:} Many recent studies have proposed and evaluated their methods on small-scale and common benchmark crowd public datasets, including UMN and UCSD~\citep{mehran2009abnormal, mahadevan2010anomaly,zhang2014social, hasan2016learning, tudor2017unmasking, cong2011sparse, alafif2021generative}.

~\citep{piciarelli2008trajectory} introduced a normal model by clustering the trajectories of moving objects for anomaly detection. Then,~\citep{mehran2009abnormal} proposed to use optical flows-based social force model to detect abnormal behaviors. A grid of particles was computed over the frames. Then, a bag of words method was applied to classify normal and abnormal behaviors. 

After~\citep{mehran2009abnormal} work,~\citep{mahadevan2010anomaly} applied learned mixtures of dynamic textures based on optical flow with salient location identification to detect abnormalities in the spatial domain. In the temporal domain, the learned mixtures of dynamic textures based on optical flow with negative log-likelihood were applied to detect abnormalities. Then,~\citep{cong2011sparse} applied a sparse reconstruction cost and a dictionary to measure normal and abnormal behaviors.

After that,~\citep{zhang2014social} introduced a social attribute-aware force model. Using an online fusion algorithm, the social attribute-aware force maps are computed. Then, global abnormal events are detected with a bag-of-words representation and local abnormal events with an abnormal map. 


Later, ~\citep{hasan2016learning} learned semi-supervised Spatio-temporal local hand-crafted features on a convolutional autoencoder to detect abnormal patterns. Histograms of oriented gradients and histograms of optical lows were used to extract the Spatio-temporal features from raw video frames to feed the convolutional autoencoder for classification.~\citep{fradi2016crowd} applied local feature tracking to describe the movements of the crowd. They represented the crowd as an evolving graph. To analyze the crowd scene for an abnormal event, mid-level features are extracted from the graph. 

~\citep{colque2016histograms} used the histograms of magnitude, orientation, and entropy of the optical flow with the nearest neighbor search algorithm to detect the anomalous. In the training phase, they stored the histograms of each moving object as normal patterns. In the testing phase, they used the nearest neighbor search to find normal patterns to find the abnormality.

~\citep{cocsar2016toward} employed trajectory features and motion features. They used a bag-of-words representation to describe the actions. Then, they applied a clustering algorithm to perform abnormal detection in an unsupervised manner. 

Followed by~\citep{hasan2016learning},~\citep{fradi2016crowd},~\citep{colque2016histograms}, and~\citep{cocsar2016toward},~\citep{tudor2017unmasking} used a sliding window technique to obtain partial video frames. The motions and appearance features were extracted from the frames and fed to a linear binary classifier to detect normality and abnormality in behaviors.

Recently,~\citep{alafif2021generative} applied a FlowNet and UNet framework to generate normal and abnormal optical flows to detect abnormalities. However, most current existing abnormal behaviors detection methods are computationally expensive since they require modeling the appearance of the frames~\citep{mahadevan2010anomaly}, particles advection~\citep{mehran2009abnormal}, sliding windows \citep{hasan2016learning}\citep{tudor2017unmasking}, dictionaries~\citep{cong2011sparse}, hand-crafted features extractors~\citep{hasan2016learning}, and generating images~\citep{alafif2021generative}. In addition to the computational efficiency drawbacks, their approach effectiveness may decrease in large-scale crowds since they have many challenges, including partial and full occlusions, different scales, blurring, and large-number abnormal behaviors.  

\item \textbf{Large-scale crowds:} Several research works studied abnormal behaviors on large-scale crowds in~\citep{solmaz2012identifying, wang2013motion, alqaysi2013detection, bera2016realtime, zou2015detect, bera2016realtime, pennisi2016online, fradi2016crowd, wu2017crowd, alafif2021generative}. 

First,~\citep{solmaz2012identifying} introduced a linear approximation using a Jacobian matrix to identify large-scale crowd abnormal behaviors. An optical flow and a particle advection were used. Then,~\citep{wang2013motion} started to cluster crowd feature maps to analyze motion patterns. Followed by~\citep{wang2013motion},~\citep{alqaysi2013detection} applied motion history image and segmented optical flow to extracted features. Then, a histogram was used for the motion direction and magnitude to detect crowd abnormal behaviors. 

Later,~\citep{zou2015detect} detected large-scale crowd motions and trajectories using tracklets association. Similar to ~\citep{zou2015detect}, ~\citep{bera2016realtime} computed abnormal behaviors trajectories using Bayesian learning techniques. Then, ~\citep{pennisi2016online} segmented extracted features to detect crowd abnormal behaviors. In recent years, ~\citep{fradi2016crowd} and ~\citep{wu2017crowd} worked on analyzing large-scale crowd properties using visual feature descriptors. 

However, existing methods are only confined to detecting and analyzing large-scale crowds as a mass. To the best of our knowledge, no existing works have detected individuals' abnormal behaviors in large-scale crowds except the work presented in~\citep{alafif2021generative}. In comparison with the recent work in~\citep{alafif2021generative}, the proposed methods don't require generating individual abnormal behavior images. Compared to the work in~\citep{alafif2021generative}, the proposed method achieves better accuracy using the HAJJv1 dataset. 
\end{itemize}


\begin{table*}[h]
\centering
 \caption{\label{tab:datasets}A comparison of existing public abnormal behaviors datasets and the Abnormal Behaviors HAJJv2 dataset.}
 \resizebox{\textwidth}{!}{%
\begin{tabular}{lp{6cm}ccc}
 
 \hline
  \textbf{Dataset} & \textbf{Abnormal Behaviors} & \textbf{Size} & \textbf{Crowd Scale} & \textbf{Reference} \\
 \hline
UMN & Escape & 24,240 KB & Small-scale &  \citep{UMN} \\
\hline
UCSD & Non-pedestrian movements & 1.74 GB & Small-scale &  \citep{mahadevan2010anomaly}\\
\hline
HAJJv1 & Standing, sitting, sleeping, running, moving in opposite crowd direction, crossing or moving in different crowd direction, and non-pedestrian movements & 831 MB & Large-scale & 
 \citep{alafif2021generative} \\
\hline
\textbf{HAJJv2} & Standing, sitting, sleeping, running, moving in opposite crowd direction, crossing or moving in different crowd direction, and non-pedestrian movements & 831 MB & Large-scale &
\textbf{-} \\
\hline
 \end{tabular}}
\end{table*}

\section{HAJJv2 Dataset}\label{sec:Hajj}
HAJJv2 dataset is introduced due to the imbalance of training examples in each class and the absence of many annotations and labeling for individuals with abnormal behaviors in the HAJJv1 dataset~\citep{alafif2021generative}. The HAJJv2 dataset consists of 18 manually collected videos from the annual Hajj religious event. All the videos are stored in an mp4 extension. The collected videos include individuals' abnormal behaviors in massive crowds. The videos are captured from different scenes and places in the wild during the Hajj event. Five videos are captured in ``Massaa'' scene while other videos are captured in ``Jamarat'', ``Arafat'', and ``Tawaf''. These videos were recorded using high-resolution cameras. Then, the videos are cropped and split into training and testing sets. Each set contains 9 short videos. Each video in the training set lasts for 25 seconds, while each video in the testing set lasts for 20 seconds.

\begin{figure}[h]
    \centering
    \begin{subfigure}[b]{0.2\textwidth}
        \centering
        \includegraphics[width=\textwidth]{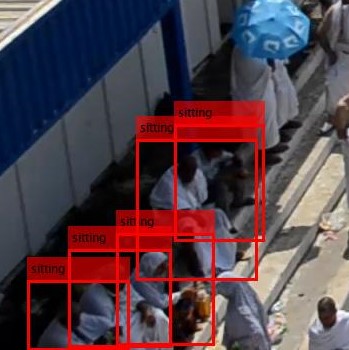}
        \caption{Sitting.}
        \label{fig:sitting}
    \end{subfigure}
    \hfill
    \begin{subfigure}[b]{0.2\textwidth}
        \centering
        \includegraphics[width=\textwidth]{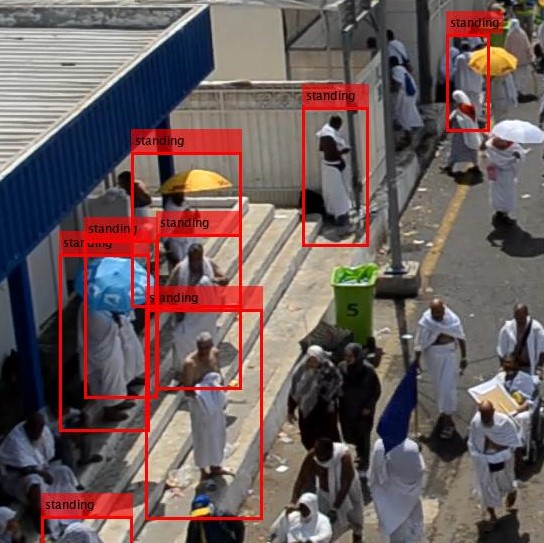}
        \caption{Standing.}
        \label{fig:standing}
    \end{subfigure}
    \hfill
        \begin{subfigure}[b]{0.2\textwidth}
        \centering
        \includegraphics[width=\textwidth]{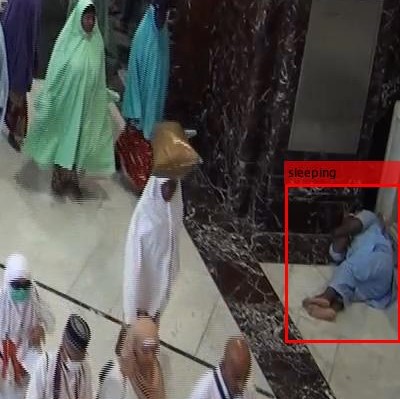}
        \caption{Sleeping.}
        \label{fig:sleeping}
    \end{subfigure}
    \hfill
    \begin{subfigure}[b]{0.2\textwidth}
        \centering
        \includegraphics[width=\textwidth]{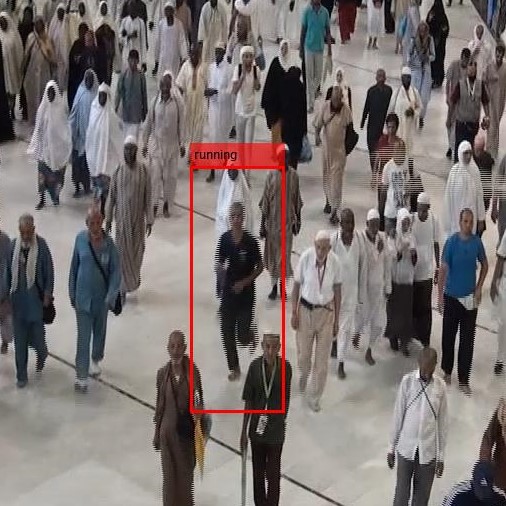}
        \caption{Running.}
        \label{fig:running}
    \end{subfigure}
    \hfill
    \begin{subfigure}[b]{0.2\textwidth}
        \centering
        \includegraphics[width=\textwidth]{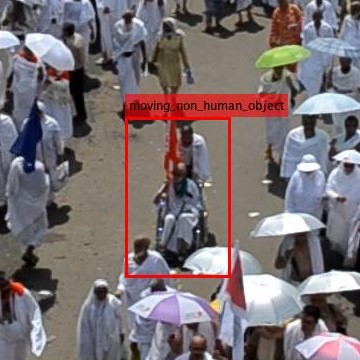}
        \caption{Moving non-pedestrian object.}
        \label{fig:moving_non_human_object}
    \end{subfigure}
    \hfill
    \begin{subfigure}[b]{0.2\textwidth}
        \centering
        \includegraphics[width=\textwidth]{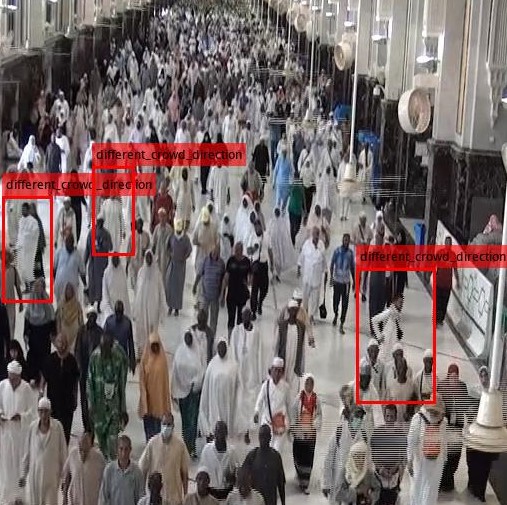}
        \caption{Moving in different direction.}
        \label{fig:different_crowd_direction}
    \end{subfigure}
    \hfill

    \begin{subfigure}[b]{0.2\textwidth}
        \centering
        \includegraphics[width=\textwidth]{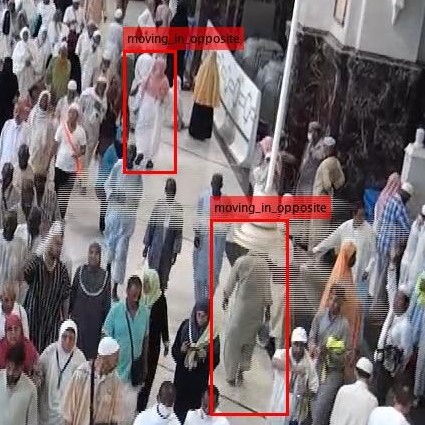}
        \caption{Moving in opposite direction.}
        \label{fig:moving_in_opposite}
    \end{subfigure}
    \hfill
    \caption{Abnormal behaviors examples in HAJJv2 dataset.}
    \label{fig:Haj}
\end{figure}

In these videos, individuals' abnormal behaviors include standing, sitting, sleeping, running, moving in opposite or different crowd directions, and non-pedestrian entities such as cars and wheelchairs. These behaviors can be potentially dangerous to large-scale crowd flows. Figure~\ref{fig:Haj} shows examples of these abnormal behaviors included in HAJJv2 dataset. The dataset statistics are provided in Table~\ref{tab:stat}. As seen in the table, the dataset is imbalanced. The sitting class has the largest number of training and testing examples, while the running class has the smallest number of examples in training and testing.

Individuals' anomalous behaviors in the videos are manually annotated and labeled for training and testing. The annotations and labeling are stored in two CSV files. The training CSV file contains 170,772 annotated and labeled individuals' abnormal behaviors, while the testing CSV file contains 129,769 annotated and labeled individuals' abnormal behaviors. A comparison of existing public abnormal behaviors datasets and the Abnormal Behaviors HAJJv2 dataset is shown in Table~\ref{tab:datasets}.


\begin{table}[h]
\centering
 \caption{\label{tab:stat} HAJJv2 dataset statistics.}
\begin{tabular}{llll}
\hline
n & Classes & Training & Testing \\
\hline
1 & Different Crowd Direction   &   7,152     &   6,262     \\
2 & Moving In Opposite          &   36,577    &   18,802    \\
3 & Moving Non Human Object     &   4,186     &   4,146     \\
4 & Running                     &   51        &   190       \\
5 & Sitting                     &   100,633   &   83,644    \\
6 & Sleeping                    &   2,400     &   2,618     \\
7 & Standing                    &   19,773    &   14,107    \\
\hline
 & Total                        &   170,772   &   129,769  \\      
\hline
\end{tabular}
\end{table}

\section{Proposed Methods}\label{sec:meth}

In this section, we present the details of our proposed methods and algorithms. First, we show the individual abnormal behaviors detection and recognition methodology pipeline and algorithm for small-scale crowds. Then, similarly, a methodology pipeline and an algorithm for detecting and recognizing abnormal behaviors are presented for large-scale crowds. Figure~\ref{fig:HiCNNRF-pipelines} shows the methodology pipelines for large-scale and small-scale crowd abnormal behavior detection and recognition.

\subsection{Individual Abnormal Behaviors Detection, Tracking, and Recognition in Small-scale Crowds}

\begin{figure*}[h]
     \centering
     \hspace{-6em}
     \begin{subfigure}[b]{0.4\textwidth}
      \centering
      \includegraphics[scale=0.048]{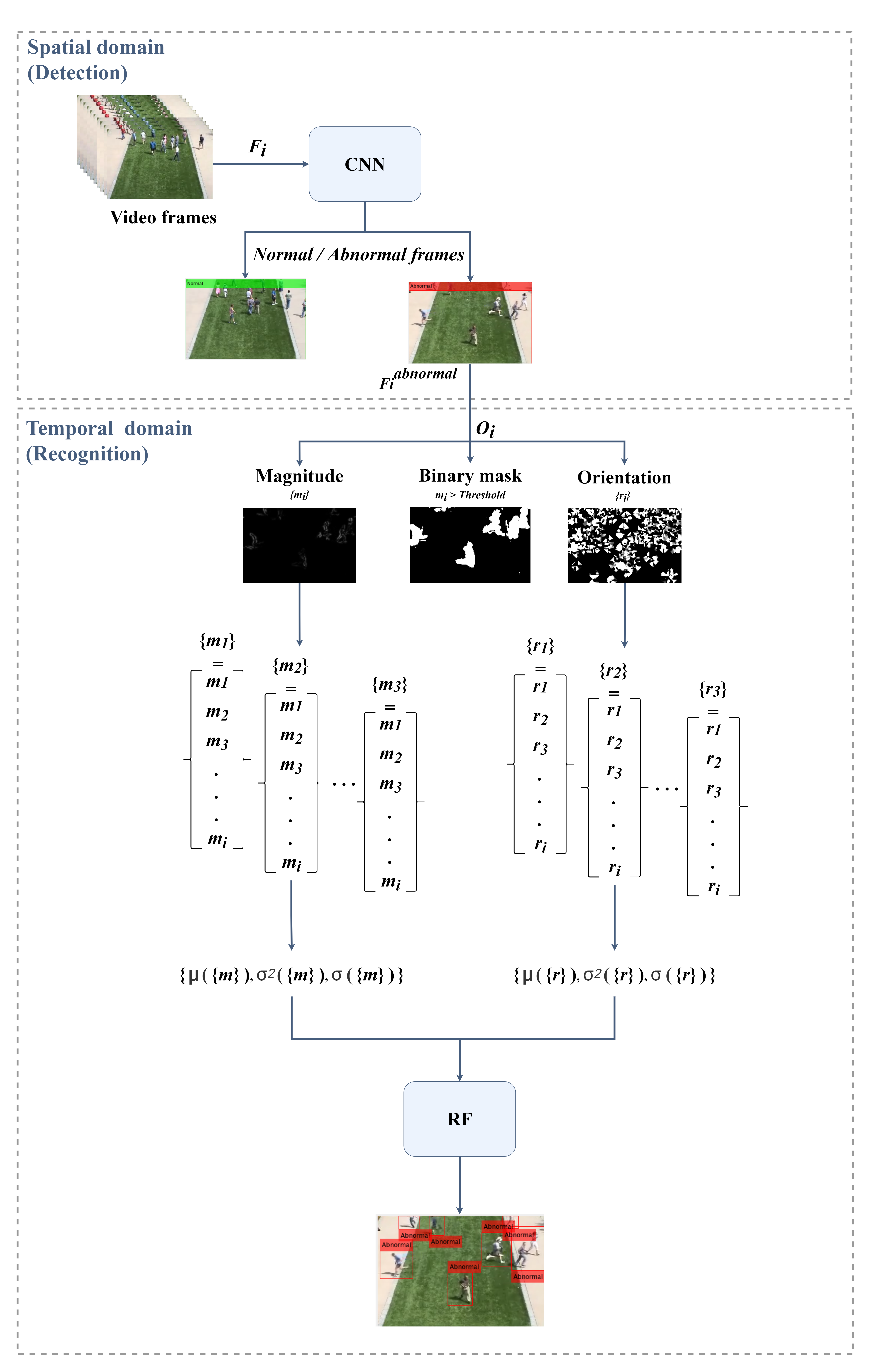}
      \caption{Small-scale crowds (UMN and UCSD datasets).}
      \label{fig:small-scale-pipeline}
     \end{subfigure}
     \hspace{6em}
     \begin{subfigure}[b]{0.4\textwidth}
          \centering
      \includegraphics[scale=0.048]{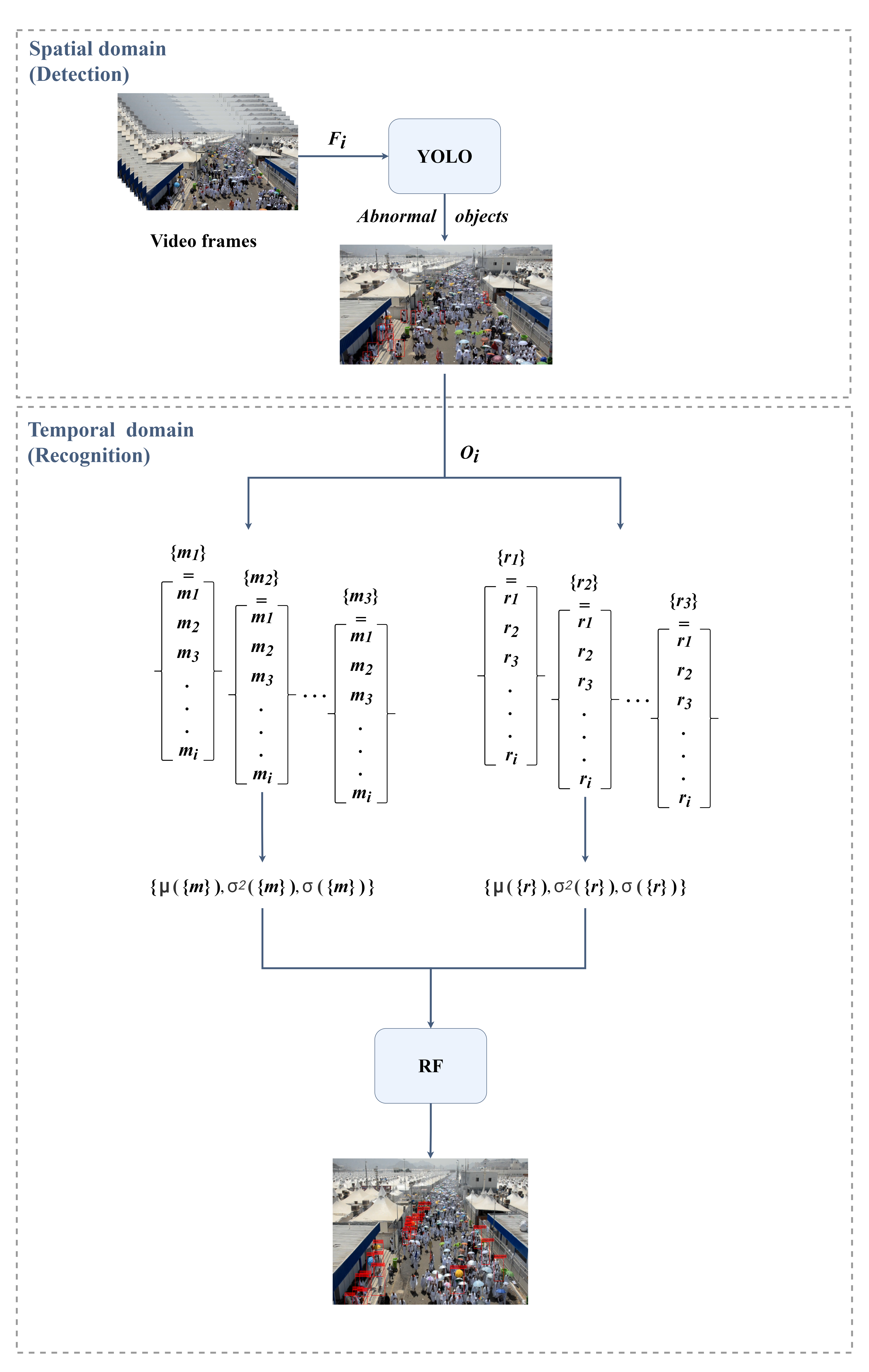}
      \caption{Large-scale crowds (HAJJv2 dataset).}
      \label{fig:Hajj-pipeline}
     \end{subfigure}
     \caption{The methodology pipelines for individuals abnormal behaviors detection, tracking, and recognition.}
     \label{fig:HiCNNRF-pipelines}
\end{figure*}

Figure~\ref{fig:small-scale-pipeline} shows the methodology pipeline for detecting and recognizing abnormal behaviors in small-scale crowded videos. The pipeline consists of spatial and temporal domains and hybrid classifiers. The spatial domain includes a pre-trained CNN classifier which focuses on classifying and detecting the abnormal behaviors generally on a frame-level. On the other hand, the temporal domain includes an RF classifier that aims to classify and recognize individuals' behaviors at an object-level within the frames.

\textit{\textbf{Spatial domain:}} Training a specialized deep model from scratch requires a vast amount of data, a lot of resources, and a long training time. Transfer learning overcomes these challenges by utilizing pre-trained deep learning models that have been trained on a huge amount of labeled data and using the previously optimized weights to perform other predictive tasks. 
Due to the lack of sufficient abnormal training datasets, we utilize transfer learning in the spatial domain. We fine-tune the pre-trained model, ResNet-50~\citep{he2016deep}, to detect 
anomalies at the frame level. Deeper networks are capable of extracting more complex feature patterns; however, they may cause a degradation problem, which degrades the detection performance. The ResNet uses a deep residual learning framework to solve the degradation problem. This gives the advantage of using a deep network to extract the complex feature patterns in the spatial domain. Therefore, we use the ResNet-50 in our experiment. 

The ResNet-50 consists of 49 convolution layers as a features extractor, followed by an average pooling and a fully connected layer as a classifier. Fine-tuning the pre-trained models is made by modifying the previous weights of the model such that they work with a new classification task. The classification layers of the pre-trained model are replaced by a fully connected layer and an output layer that outputs values equal to the number of classes. Anomalies detection is a binary classification problem. Thus, the classifier is trained on normal and abnormal frames. Therefore, we fine-tune the ResNet-50 as a binary classifier using video frames from small-scale crowd datasets. We replace the last layer with a fully connected layer that maps 2,048 units into 128, followed by an output layer that maps 128 units into 2 units representing the normal and the abnormal probabilities. Since the ResNet-50 processes inputs with a size 224$\times$224$\times$3, we resize the frames to ResNet-50's input size. A feed-forward and backpropagation algorithm is applied by updating the errors and weights to converge.

Figure~\ref{fig:small-scale-pipeline} shows detected normal and abnormal frames resulted from the ResNet-50 classifier. The detected normal frames appear in green while the detected abnormal ones appear in red.

\begin{equation} \label{eq2}
\begin{split}
\mu(m^j) = \frac{\sum_{i=1}^{p}(m_{i}^j)}{p}\\
\mu(r^j) = \frac{\sum_{i=1}^{p}(r_{i}^j)}{p}\\
\sigma^2(m^j) = \frac{\sum_{i=1}^{p}((m_{i}^j - \mu(m^j))^2)}{p}\\ \sigma^2(r^j) = \frac{\sum_{i=1}^{p}((r_{i}^j - \mu(r^j))^2)}{p}\\
\sigma(m^j) = \sqrt{\frac{\sum_{i=1}^{p}((m_{i}^j - \mu(m^j))^2)}{p}}\\ 
\sigma(r^j) =\sqrt{\frac{\sum_{i=1}^{p}((r_{i}^j - \mu(r^j))^2)}{p}}\\
\end{split}
\end{equation}

\begin{figure}[h]
    \centering
    \begin{subfigure}[b]{0.2\textwidth}
        \centering
        \includegraphics[width=\textwidth]{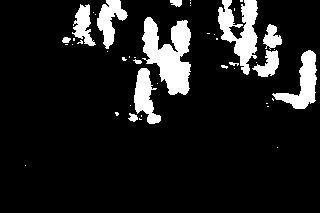}
        \caption{UMN scene 1.}
        \label{fig:maskumn1}
    \end{subfigure}
    \hfill
    \begin{subfigure}[b]{0.2\textwidth}
        \centering
        \includegraphics[width=\textwidth]{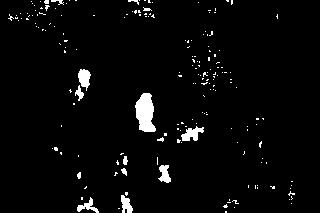}
        \caption{UMN scene 2.}
        \label{fig:maskumn2}
    \end{subfigure}
    \hfill
    \begin{subfigure}[b]{0.2\textwidth}
        \centering
        \includegraphics[width=\textwidth]{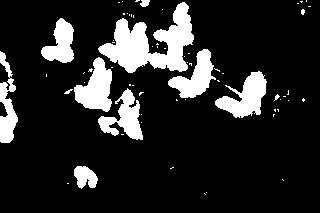}
        \caption{UMN scene 3.}
        \label{fig:maskumn3}
    \end{subfigure}
    \hfill
    \begin{subfigure}[b]{0.2\textwidth}
        \centering
        \includegraphics[width=\textwidth]{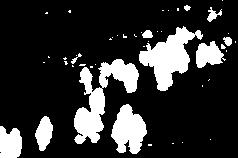}
        \caption{UCSD Ped1.}
        \label{fig:maskped1}
    \end{subfigure}
    \hfill
        \begin{subfigure}[b]{0.2\textwidth}
        \centering
        \includegraphics[width=\textwidth]{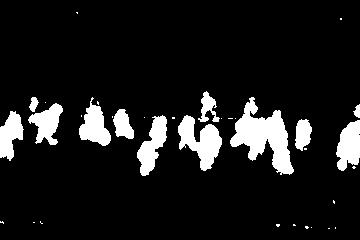}
        \caption{UCSD Ped2.}
        \label{fig:maskped2}
    \end{subfigure}
    \hfill
    \caption{Examples of binary magnitude-based masks on small-scale datasets.}
    \label{fig:masks}
\end{figure}

\textit{\textbf{Temporal domain:}}
We use the optical flow to detect the anomalies on a pixel level. By analyzing the optical flow, we can observe crowd movements, instantaneous velocity, orientations, and magnitudes. These low-level features are used to recognize individuals' behaviors. After detecting the anomalies at $i$-th frame ($F_i$), the optical flow of this frame ($O_i$) is computed using Horn-Schunck optical flows~\citep{horn1981determining}. Then, magnitudes ($m_{i}$) and orientations ($r_{i}$) features are automatically extracted from the optical flows. In small-scale crowded videos, binary magnitude-based masks using a threshold ($T$) are initiated to localize and track individuals within the frame (i.e., $\{I_{j}^i\}_{j=1}^{z}$ is the $j$-th individual in the $i$-th frame). Figure~\ref{fig:masks} shows the binary magnitude-based mask of the small-scale datasets. 
After extracting the magnitudes and orientations features, the statistical features including means ($\mu$), variances ($\sigma^2$), and standard deviations ($\sigma$) are computed. They are computed for the total pixels ($p$) of the area that represents an individual ($j$), for both the magnitudes ($m^j$) and orientations ($r^j$) as follows:

Then, these statistical features are fed to the RF classifier for training to classify and recognize temporal individual abnormal behaviors. Algorithm~\ref{alg:alg1} shows the computational steps of the proposed method in the small-scale crowds. The algorithm runs $O(n^2)$ in the worst case.

\subsection{Individual Abnormal Behaviors Detection and Recognition in Large-scale Crowds}


Figure~\ref{fig:Hajj-pipeline} shows the methodology pipeline for detecting and recognizing abnormal behaviors in large-scale crowd scenes in the HAJJv2 dataset. Similar to the method presented for small-scale crowd scenes, the method for large-scale crowded scenes also consists of hybrid classifiers. The first classifier is accountable for detecting the abnormal behavior frames in the spatial domain, while the second classifier is employed for recognizing the individuals' abnormal behaviors in the temporal domain.

\textit{\textbf{Spatial domain:}} Similar to the previous detection method on small-scale crowds, we fine-tune another pre-trained CNN model, ResNet-50. We train the ResNet-50 as a one-class classifier using all abnormal behaviors in the training set of the HAJJv2 dataset. The main goal of the ResNet-50 model is to only detect individuals' abnormal behaviors in the frames if they exist. To address the problem of the overlapped white areas when a large number of individuals and a large number of partial occlusions occur, YOLOv2~\citep{redmon2017yolo9000} technique is employed to locate individuals with abnormal behaviors in the spatial domain. We use the backpropagation algorithm to update the errors and weights in the ResNet-50 model until convergence. 

\textit{\textbf{Temporal domain:}} After detecting all individuals with abnormal behaviors, we employ Horn-Schunck optical flows ($O_{i}$) on the detected individuals. This approach is different from the previous method applied for small-scale crowds. We extract the $m_{i}$ and the $r_{i}$ features from the resulted optical flows.
To track individuals with abnormal behaviors, a Kalman filter~\citep{welch1995introduction} is used directly with 
YOLOv2 detector. The Kalman filter predicts individuals' locations in the next frames. We avoid using the binary magnitude-based masks since they mainly cause overlapping contiguous groups of white pixels due to heavy partial occlusions in large-scale crowded scenes.

After detecting individuals' abnormal behaviors and extracting their statistical features, we compute the means ($\mu$), variances ($\sigma^2$), and standard deviations ($\sigma$) for each individual similar to the small-scale crowd method. Then, an RF classifier is used as a multi-class classifier to classify and recognize all individuals with abnormal behaviors. 

Algorithm~\ref{alg:alg1} shows the sequence of our implementable method. Similar to algorithm~\ref{alg:alg1}, algorithm~\ref{alg:alg2} also runs $O(n^2)$ in the worst case.

\begin{algorithm}[h]
    \SetKwInOut{Input}{Input}
    \SetKwInOut{Output}{Output}
    \Input{Video frame sequences $\{F_{i}\}_{i=1}^{n}$, where $F_{i}$ consists of a number of frames $f$ such that $F_{i}=\lbrace f_{1}, f_{2},\dots,f_{n}\rbrace$.}
   \Output{Abnormal behavior frames and objects.}
   Use $F_{i}$ to fine-tune a pre-trained CNN model in the spatial domain using feed-forward and back-propagation algorithm until convergence and update the weights\;
   Compute optical flow $\{O_{i}\}_{i=1}^{n-1}$ from the original video sequence $F_{i}$ : $F_{i} \longrightarrow \lbrace O_{1}, O_{2},\dots, O_{n-1} \rbrace$\; 
   Extract optical flows orientations $\{r_{i}\}_{i=1}^{n}$  and magnitudes $\{m_{i}\}_{i=1}^{n}$ features from $O_{i}$ : $O_{i}= \lbrace (r_{1}, m_{1}), (r_{2}, m_{2}),\dots, (r_{n-1}, m_{n-1}) \rbrace $\;
  Create the binary magnitude-based mask using a threshold $T$, $mask = m_i > T$\;
   Extract the individuals within the mask $\{I_{j}^i\}_{j=1}^{z}$\;
   Compute orientations and magnitudes means $\{\mu_{j}^i\}_{j=1}^{z}$, variances $\{\sigma_{j}^i\}_{j=1}^{z}$, and standard deviations $\{{\sigma^2}_{j}^i\}_{j=1}^{z}$\;
   Use the statistical features $\{\mu_{j}^i\}_{j=1}^{z}$, $\{\sigma_{j}^i\}_{j=1}^{z}$, and $\{{\sigma^2}_{j}^i\}_{j=1}^{z}$ of the individuals to train temporal normal and abnormal behaviors using an RF model\;
   Insert a test video frame sequences $\{F_{i}^t\}_{i=1}^{n}$\;
   \While{$F_{i}^t \neq$ empty}{
   Use $F_{i}^t$ to test the fine-tuned CNN model\;
    \If{$F_{i}^t$ is abnormal}
      {
        Compute optical flows $O_{i}^t$\;
        Extract optical flows orientations $r_{i}^t$ and magnitudes $m_{i}^t$ features from $O_{i}^t$\;
        Create binary magnitude-based mask to localize and track individuals $\{I_{j}^{t,i}\}_{j=1}^{z}$\;
        \While {$\{I_{j}\}^{t,i} \neq$ empty}{
        Test the statistical features: means $\mu_{j}^{t,i}$, variances $\sigma_{j}^{t,i}$, and standard deviations ${\sigma^2}_{j}^{t,i}$ features to classify $I_{j}^{t,i}$ using the trained RF model\;
        }
      }
      }
    \caption{A hybrid CNN and RF algorithm for Spatio-temporal small-scale crowd abnormal behaviors detection, tracking, and recognition in a video.}
    \label{alg:alg1}
\end{algorithm}

\begin{algorithm}[h]
    \SetKwInOut{Input}{Input}
    \SetKwInOut{Output}{Output}

    \Input{Video frame sequence $\{AB_{i}\}_{i=1}^{n}$, where $AB_{i}$ consists of a number of abnormal behaviors examples $ab$ such that $AB_{i}= \lbrace ab_{1}, ab_{2},\dots, ab_{n} \rbrace $.}
   \Output{Abnormal behavior objects.}
    Use $AB_{i}$ as one class to fine-tune a pre-trained CNN model in the spatial domain using feed-forward and back-propagation algorithm until convergence and update the weights\;
    Compute the optical flow $\{O_{i}\}_{i=1}^{n-1}$ from the original video sequence $AB_i$ : $AB_i \longrightarrow \{O_1, O_2, \dots, O_{n-1}\}$\;
    Extract optical flows orientations $\{r_i\}_{i=1}^{n-1}$ and magnitudes $\{m_i\}_{i=1}^{n-1}$ from $O_{i}$ : $O_{i}= \lbrace (r_{1}, m_{1}), (r_{2}, m_{2}),\dots, (r_{n-1}, m_{n-1}) \rbrace $\;
    Compute the orientations and magnitudes means $\{\mu_{j}^i\}_{j=1}^{z}$, variances $\{\sigma_{j}^i\}_{j=1}^{z}$, and standard deviations $\{{\sigma^2}_{j}^i\}_{j=1}^{z}$ for each abnormal behaviors examples $ab_j$ within the frame $AB_i$\;
    Train multi-class temporal abnormal behaviors features using an RF model\;
   Insert a test video frame sequences $\{AB_{i}^t\}_{i=1}^{n}$\;
   \While{$\{AB_{i}^t\} \neq$ empty}{
   Use $AB_{i}^t$ to test the fine-tuned CNN model\;
   Compute the optical flow $O_{i}^t$\;
    \While{$\{ab_j^{i,t}\}_{j=1}^{z} \neq $empty} 
      {
        Compute the orientations and magnitudes means  $\mu_{j}^{i,t}$, variances $\sigma_{j}^{i,t}$, and standard deviations ${\sigma^2}_{j}^{i,t}$ for $ab_{j}^{i,t}$\;
        Use $\{\mu_{j}^{i,t}, \sigma_{j}^{i,t}, {\sigma^2}_{j}^{i,t}\}$ to test the trained RF model\;
        Use Kalman filter to track $ab_{j}^{i,t}$\; 
      }
      }
    \caption{A hybrid CNN and RF algorithm for Spatio-temporal large-scale crowd abnormal behaviors detection, tracking, and recognition in a video.}
    \label{alg:alg2}
\end{algorithm}
\vspace{-5em}

\section{Experiments}\label{sec:Exp}
In this section, we first provide details of the implementation of the proposed method. Second, we briefly describe the benchmark datasets used in the experiment. Third, we show our abnormal behavior detection and recognition qualitative and quantitative results. Then, we compare the results with the existing and the most recent methods for abnormal behavior detection in small-scale and large-scale crowds. 

  
\subsection{Implementation}
We implemented the proposed methods in MATLAB R2020b. The ResNet-50 and the RF models were trained using an NVIDIA Tesla V100S GPU server with 32 GB of RAM. 

 
\subsection{Datasets}
In this section, we use public datasets such as UMN~\citep{UMN}, UCSD~\citep{mahadevan2010anomaly}, HAJJv1~\citep{alafif2021generative}, and HAJJv2 datasets to evaluate the proposed method on small-scale and large-scale crowds. HAJJv2 is described in Section~\ref{sec:Hajj}. The UMN and USCD used datasets are summarized as follows: \\

\begin{figure}[h]
     \centering
          \begin{subfigure}[b]{0.2\textwidth}
         \centering
         \includegraphics[width=\textwidth]{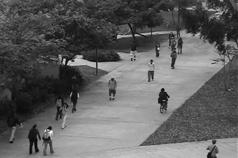}
         \caption{Abnormal Ped1.}
         \label{fig:ped1-ab}
     \end{subfigure}
     \hfill
     \begin{subfigure}[b]{0.2\textwidth}
         \centering
         \includegraphics[width=\textwidth]{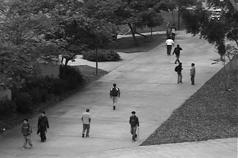}
         \caption{Normal Ped1.}
         \label{fig:ped1-n}
     \end{subfigure}
     \hfil
     \begin{subfigure}[b]{0.2\textwidth}
         \centering
         \includegraphics[width=\textwidth]{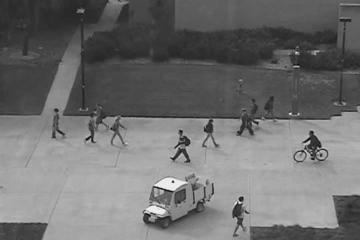}
         \caption{Abnormal Ped2.}
         \label{fig:ped2-ab}
     \end{subfigure}
     \hfill
     \begin{subfigure}[b]{0.2\textwidth}
         \centering
         \includegraphics[width=\textwidth]{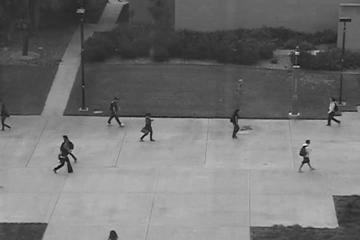}
         \caption{Normal Ped2.}
         \label{fig:ped2-n}
     \end{subfigure}
     \hfill
        \caption{Examples of normal and abnormal behaviors in UCSD dataset from two different outdoor scenes.}
        \label{fig:ucsd-Frames}
\end{figure}

\begin{itemize}
\item \textbf{The University of Minnesota (UMN) dataset.} The UMN dataset is a small-scale crowd dataset that contains three different unrealistic scenes. Two scenes were recorded outdoor while one was recorded indoor. Each UMN scene starts with a normal activity followed by abnormal behavior. Walking, for example, is considered normal activity, while running is an abnormal one. The frame resolution in UMN scenes is 320 x 240 pixels. The abnormal frames contain a short description at the top of the frames. Thus, we apply a pre-processing technique on the frames to remove the pixels that contain these descriptions to avoid biases in training and testing the model in the experiment. Figure~\ref{fig:UMN-Frames} illustrates an example of UMN's frames. The train and test splits are not explicitly specified. Moreover, the annotations are only available at the frame level. Because of these ambiguities, we use 70\% of the frames for training and the rest for testing. To address the lack of pixel-level annotations, we consider all objects in the abnormal frames as abnormal individuals and all objects in the normal frames as normal individuals. The UMN scenes are evaluated separately since they have illumination and background variations.

\begin{figure}[h]
     \centering
          \begin{subfigure}[b]{0.2\textwidth}
         \centering
         \includegraphics[width=\textwidth]{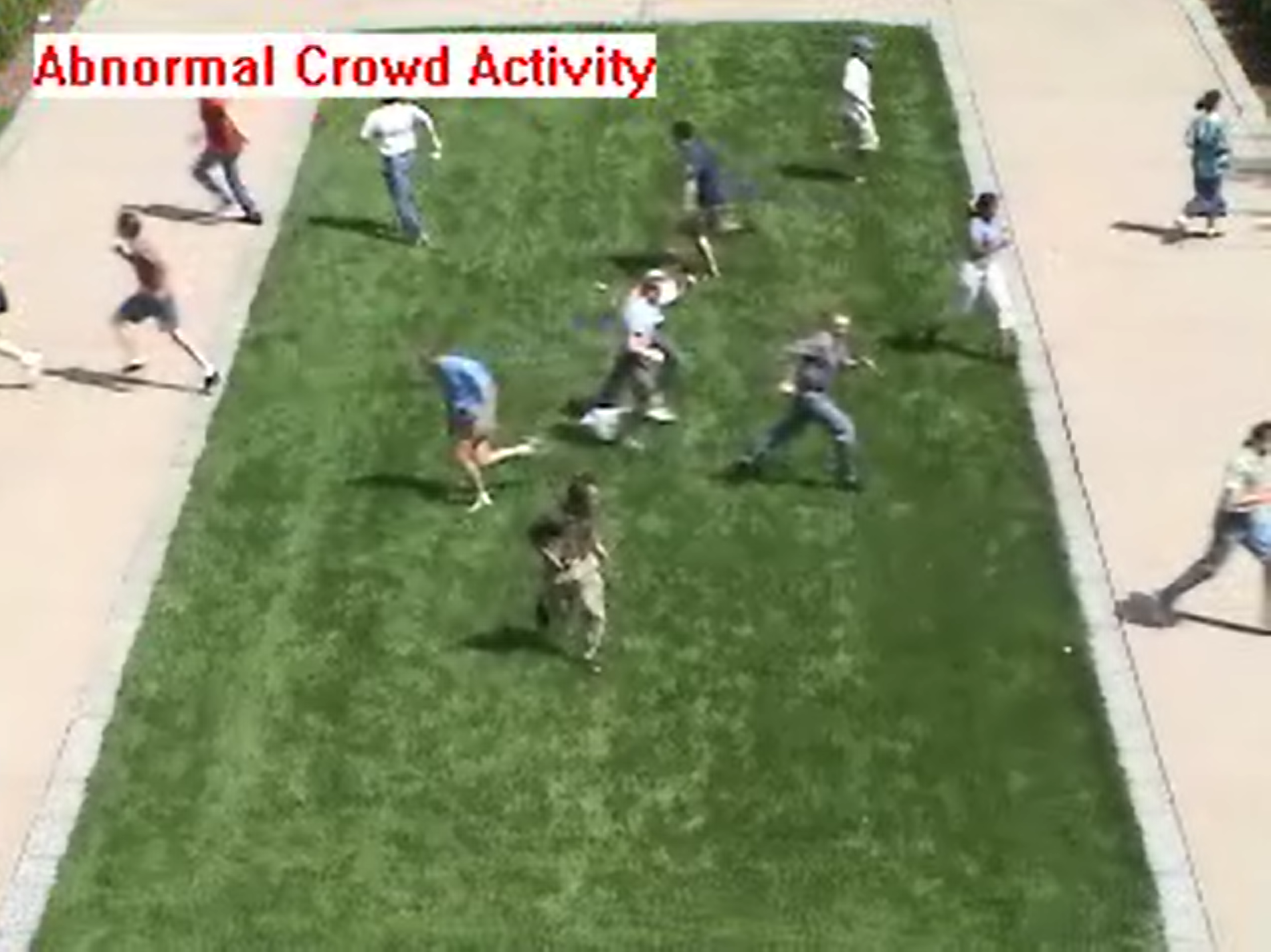}
         \caption{Outdoor abnormal behaviors in scene 1. }
         \label{fig:umn-outdoor2-abnormal}
     \end{subfigure}
     \hfill
     \begin{subfigure}[b]{0.2\textwidth}
         \centering
         \includegraphics[width=\textwidth]{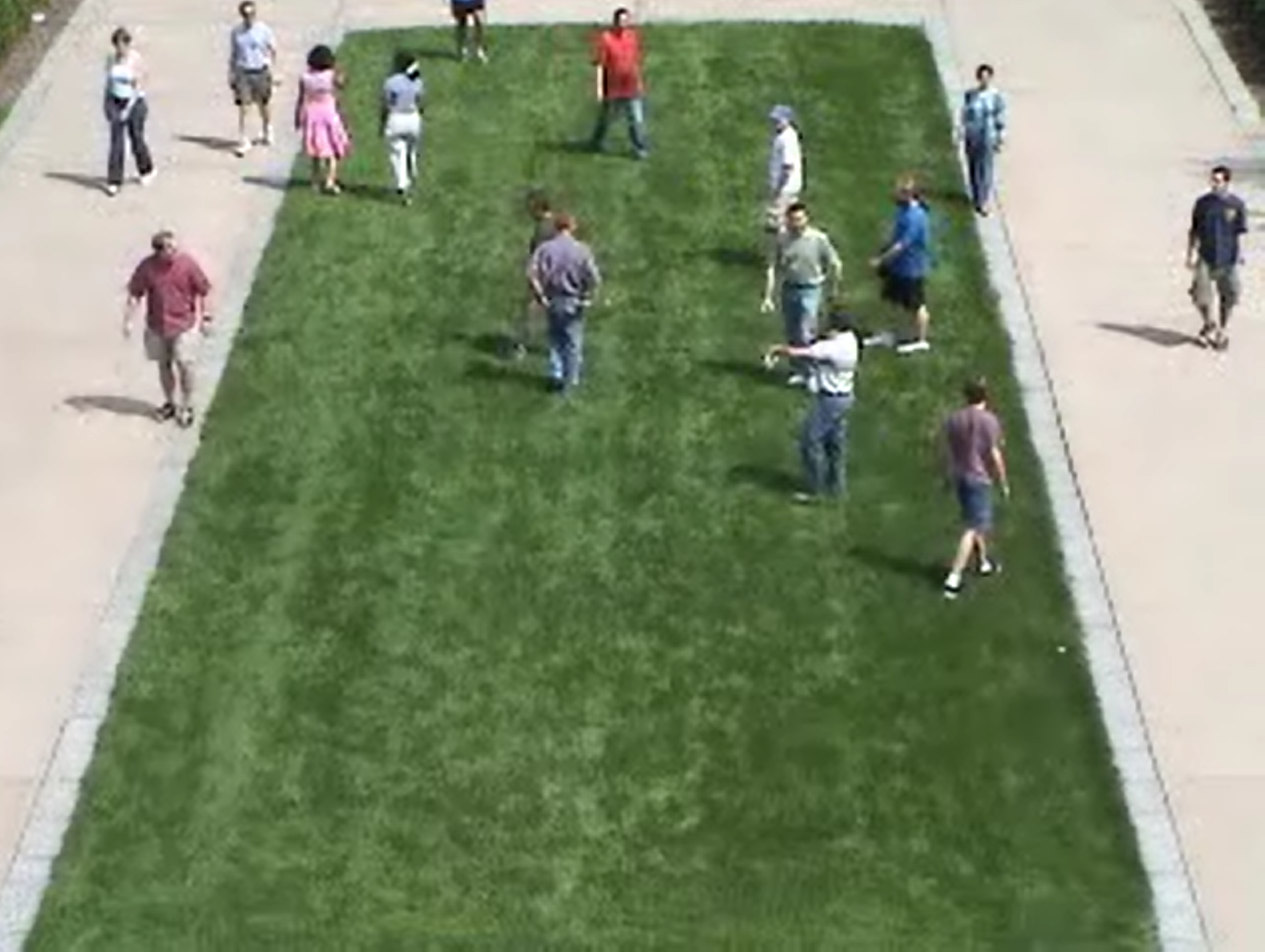}
         \caption{Outdoor normal behaviors in scene 1.}
         \label{fig:umn-outdoor2-normal}
     \end{subfigure}
     \hfill
     \begin{subfigure}[b]{0.2\textwidth}
         \centering
         \includegraphics[width=\textwidth]{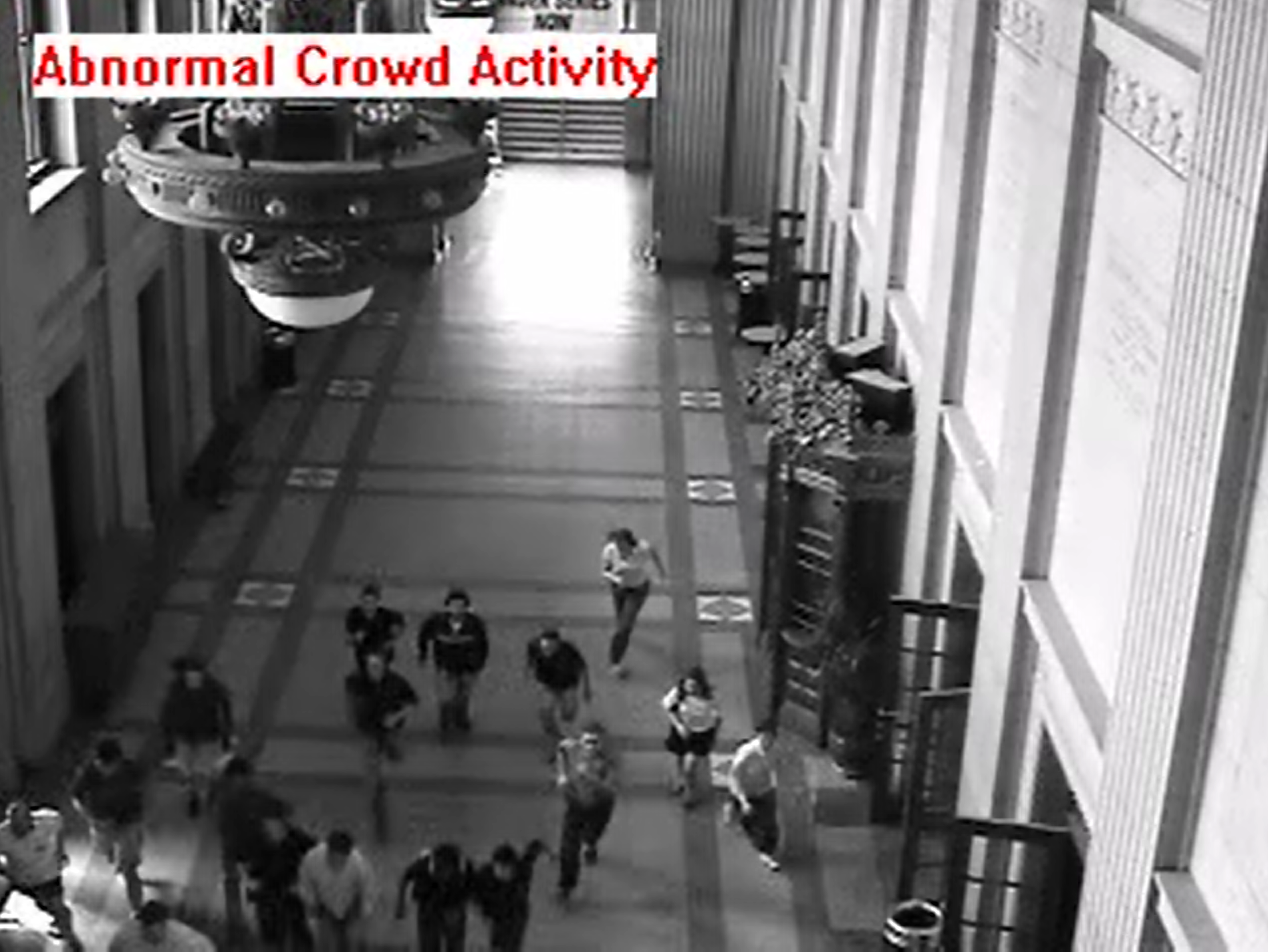}
         \caption{Indoor abnormal behaviors in scene 2.}
         \label{fig:umn-indoor-abnoraml}
     \end{subfigure}
     \hfill
     \begin{subfigure}[b]{0.2\textwidth}
         \centering
         \includegraphics[width=\textwidth]{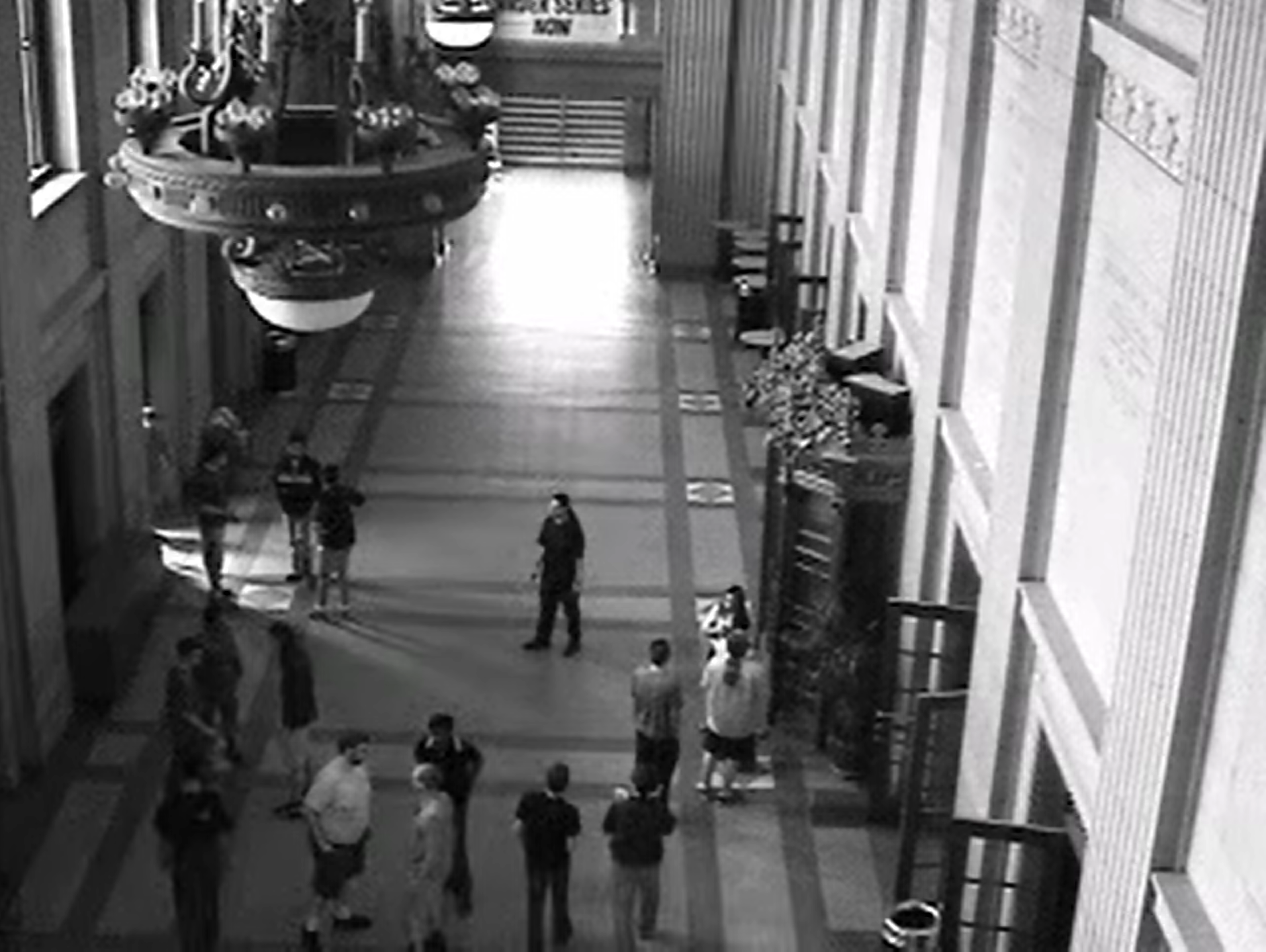}
         \caption{Indoor normal behaviors in scene 2.}
         \label{fig:umn-indoor-normal}
     \end{subfigure}
     \hfill
     \begin{subfigure}[b]{0.2\textwidth}
         \centering
         \includegraphics[width=\textwidth]{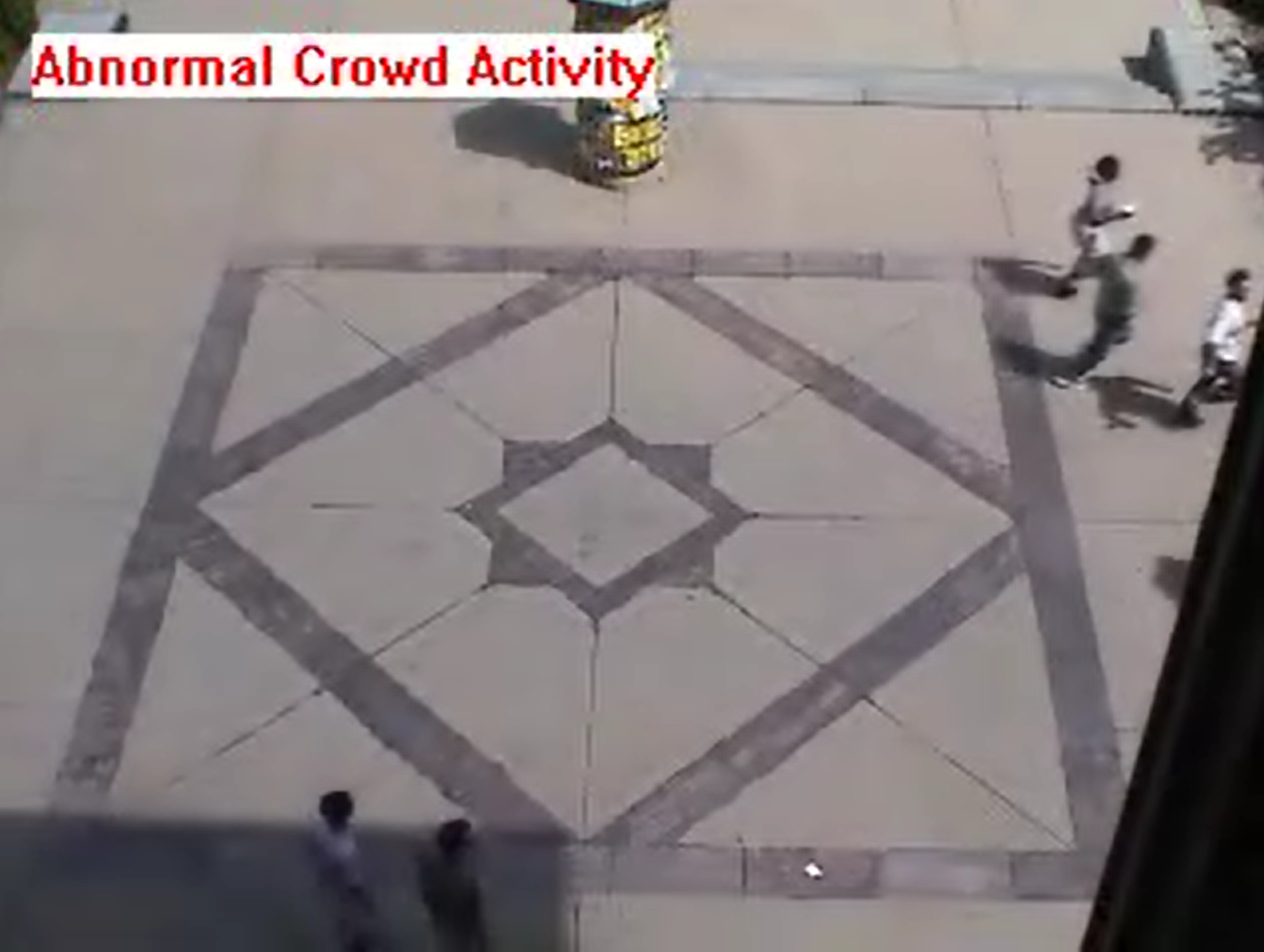}
         \caption{Outdoor abnormal behaviors in scene 3.}
         \label{fig:umn-outdoor1-abnormal}
     \end{subfigure}
     \hfill
     \begin{subfigure}[b]{0.2\textwidth}
         \centering
         \includegraphics[width=\textwidth]{UMN_Frames_Ex1.png}
         \caption{Outdoor normal behaviors in scene 3.}
         \label{fig:umn-outdoor1-normal}
     \end{subfigure}
     \hfill
        \caption{Examples of normal and abnormal behaviors in UMN dataset from three different indoor and outdoor scenes.}
        \label{fig:UMN-Frames}
\end{figure}

\item \textbf{The University of California, San Diego (UCSD) dataset.} The UCSD dataset is also a small-scale crowd dataset that consists of two subsets, namely Pedestrian 1 (Ped1) and Pedestrian 2 (Ped2). The dataset contains clips from independent static cameras viewing pedestrian walkways. It includes abnormal behaviors such as bicycles, cars, carts, skateboards, and wheelchairs as non-pedestrian objects. Ped1 contains 34 normal behavior videos and 16 abnormal behavior videos. Each video contains 200 frames with a resolution of 238 x 158 pixels. Ped2 contains 16 normal behavior videos and 12 abnormal behavior videos. The videos have different numbers of frames with a resolution of 360 x 240 pixels. Both temporal and spatial annotations are provided. Thus, the UCSD is appropriate for locating and tracking abnormal objects in small-scale crowds. In our experiment, we used both normal and abnormal videos for training and testing. Figure~\ref{fig:ucsd-Frames} illustrates some examples from Ped1 and Ped2 frames. 
\end{itemize}

\begin{figure}[h]
    \centering
    \begin{subfigure}[b]{0.2\textwidth}
        \centering
        \includegraphics[width=\textwidth]{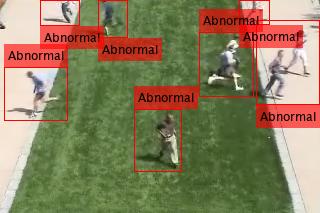}
        \caption{UMN scene 1.}
        \label{fig:umn1-Results}
    \end{subfigure}
    \hfill
    \begin{subfigure}[b]{0.2\textwidth}
        \centering
        \includegraphics[width=\textwidth]{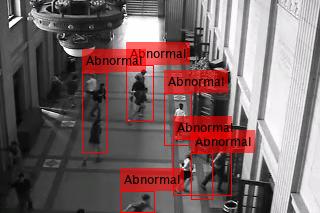}
        \caption{UMN scene 2.}
        \label{fig:umn2-Results}
    \end{subfigure}
    \hfill
    \begin{subfigure}[b]{0.2\textwidth}
        \centering
        \includegraphics[width=\textwidth]{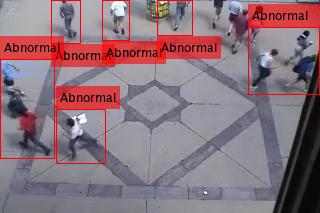}
        \caption{UMN scene 3.}
        \label{fig:umn3-Results}
    \end{subfigure}
    \hfill
    \begin{subfigure}[b]{0.2\textwidth}
        \centering
        \includegraphics[width=\textwidth]{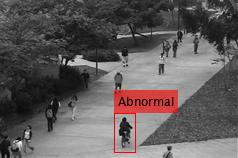}
        \caption{UCSD Ped1.}
        \label{fig:ped2-Results}
    \end{subfigure}
    \hfill
        \begin{subfigure}[b]{0.2\textwidth}
        \centering
        \includegraphics[width=\textwidth]{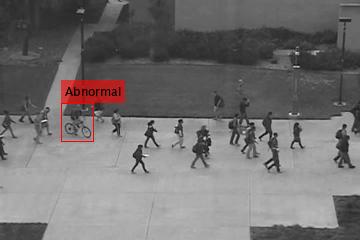}
        \caption{UCSD Ped2.}
        \label{fig:ped2-Results}
    \end{subfigure}
    \hfill
    \caption{Our qualitative results on small-scale crowds datasets.}
    \label{fig:qualitative-small}
\end{figure}

\begin{figure}[h]
    \centering
    \begin{subfigure}[b]{0.2\textwidth}
        \centering
        \includegraphics[width=\textwidth]{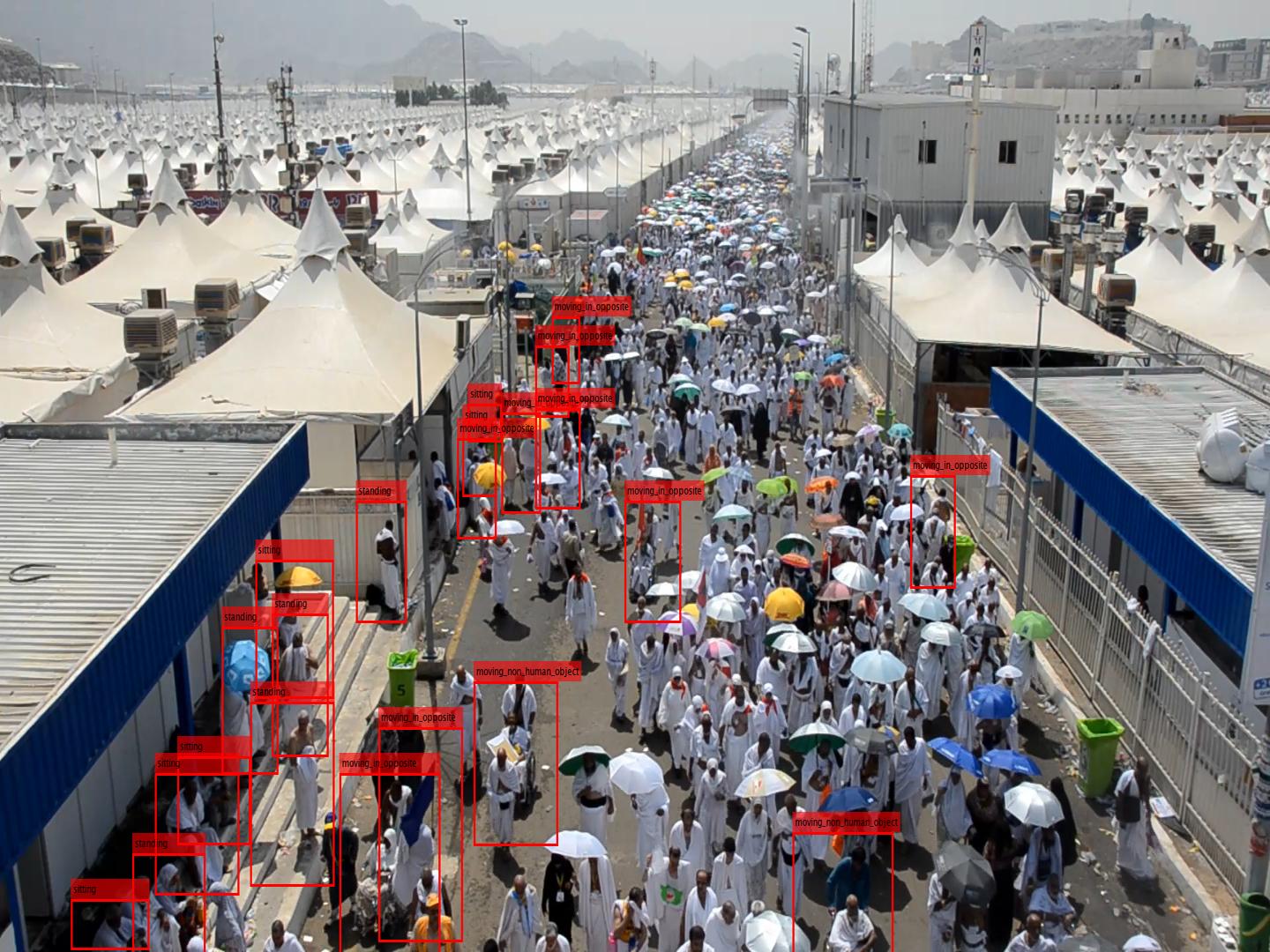}
        \caption{Video No. 10 from Arafat scene.}
        \label{fig:HAJJ-Results-v10}
    \end{subfigure}
    \hfill
    \begin{subfigure}[b]{0.2\textwidth}
        \centering
        \includegraphics[width=\textwidth]{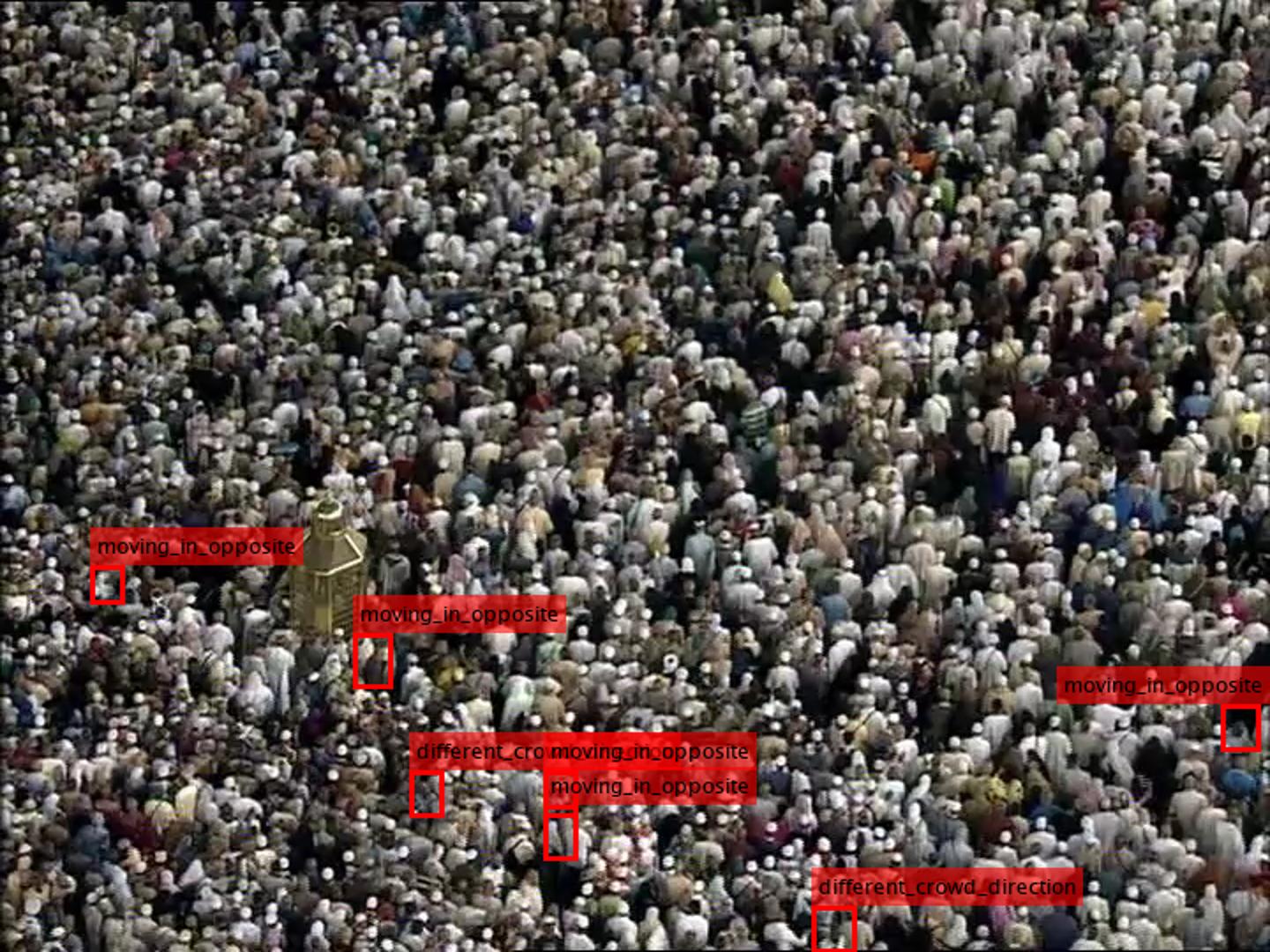}
        \caption{Video No. 12 from Tawaf scene.}
        \label{fig:HAJJ-Results-v12}
    \end{subfigure}
    \hfill
    \begin{subfigure}[b]{0.2\textwidth}
        \centering
        \includegraphics[width=\textwidth]{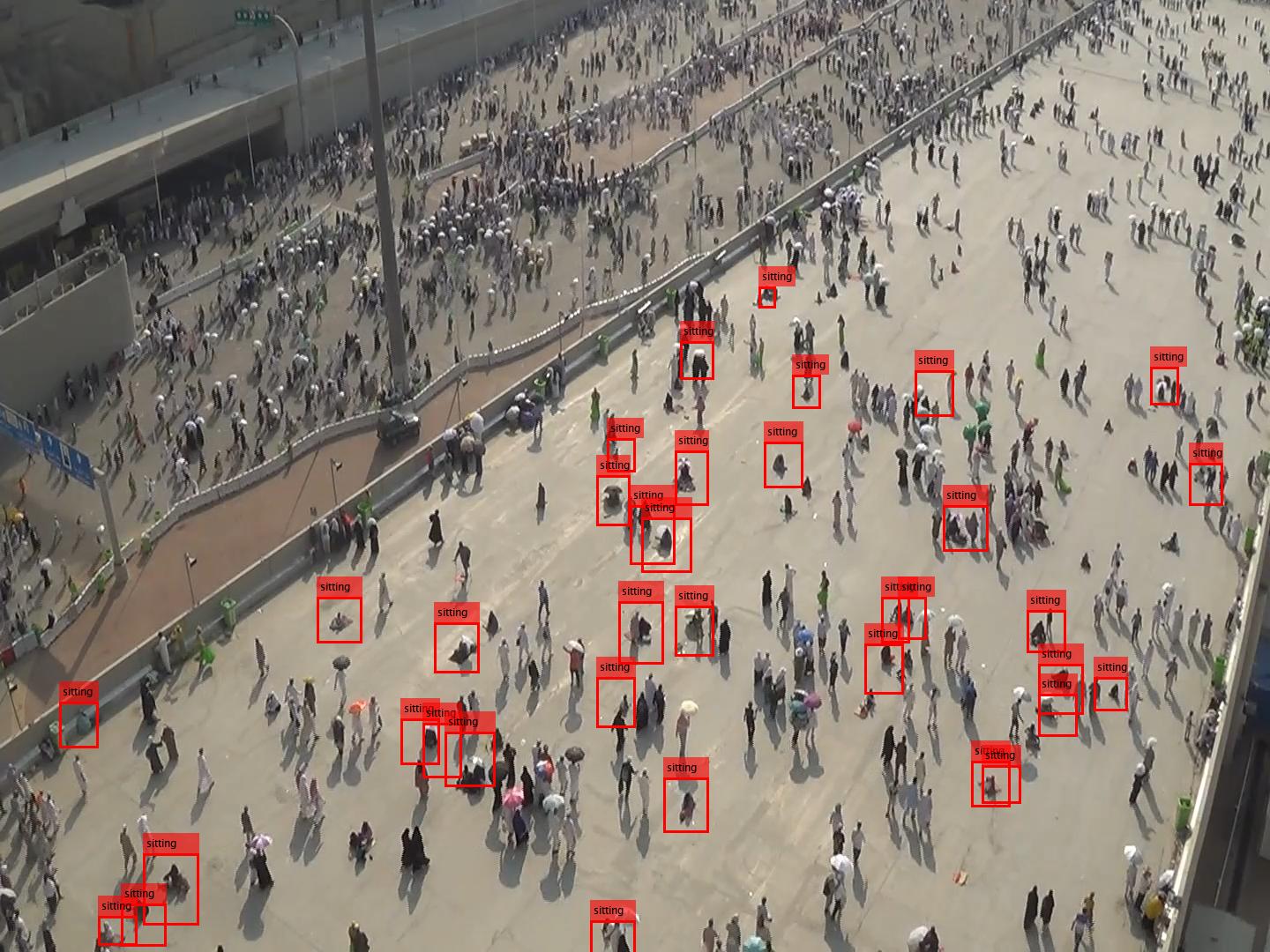}
        \caption{Video No. 9 from Jamarat scene.}
        \label{fig:HAJJ-Results-v9}
    \end{subfigure}
    \hfill
    \begin{subfigure}[b]{0.2\textwidth}
        \centering
        \includegraphics[width=\textwidth]{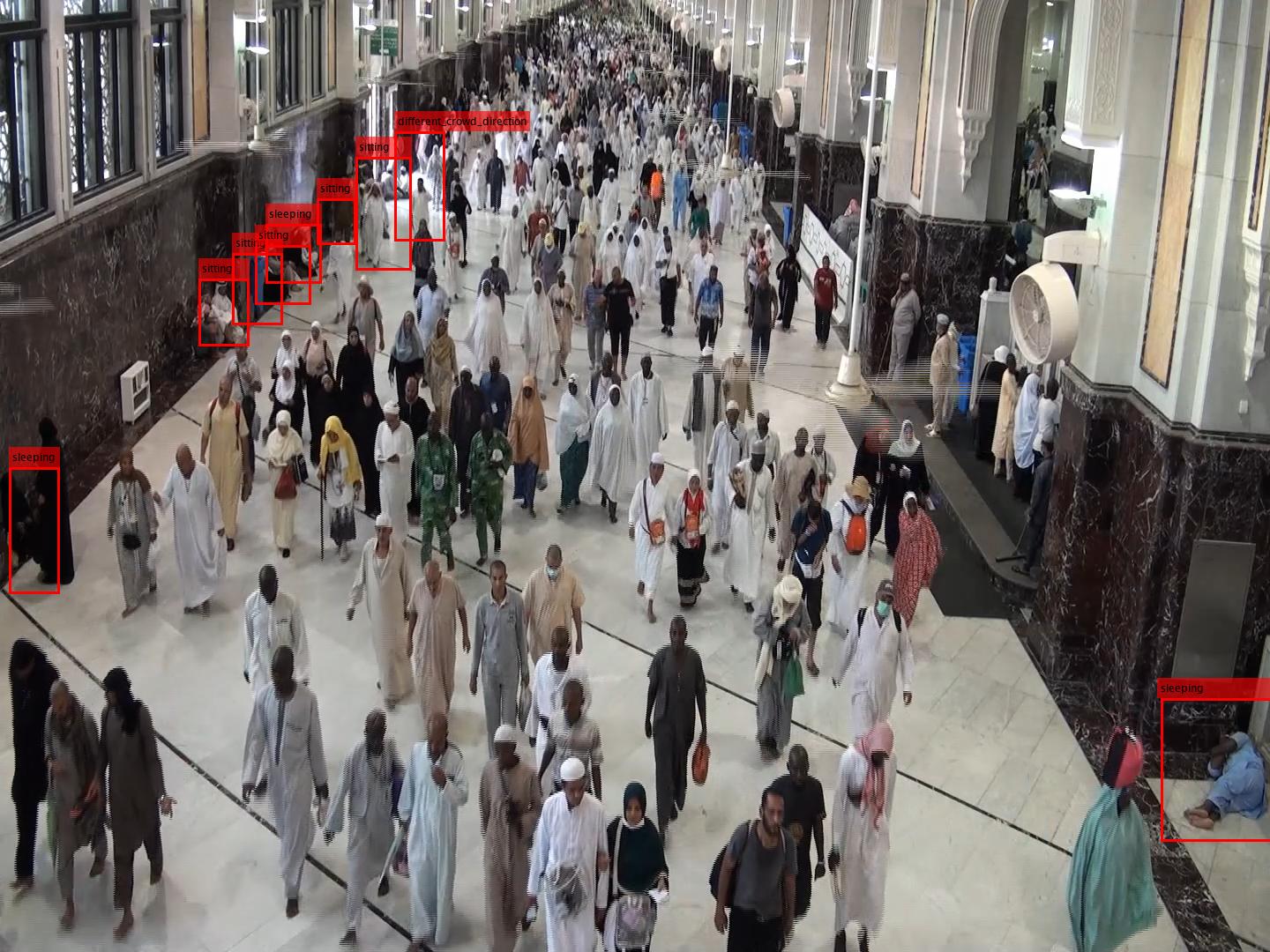}
        \caption{Video No. 5 from Masaa scene.}
        \label{fig:HAJJ-Results-v5}
    \end{subfigure}
    \hfill
    \caption{Our qualitative results on large-scale crowds dataset using HAJJv2.}
    \label{fig:qualitative-Hajj}
\end{figure}

\subsection{Experimental Settings and Hyperparameters}
For both small-scale and large-scale crowds experiments, different configurations are evaluated to determine the most effective approach. Details are described in the following.\\

\textbf{Small-scale crowds:} Different pre-trained CNN models such as ResNet-50, VGG-16, VGG-19, AlexNet, and SqueezeNet were examined in the spatial domain model as a part of the proposed method. According to our preliminary experiments, the ResNet-50 model achieves better performance on small-scale crowd datasets. We fine-tune the ResNet-50 model using Adam optimizer~\citep{kingma2014adam} with a learning rate of 0.0001 for 15 epochs and 128 normal and abnormal frames per batch of each dataset. 


Many methods to estimate optical flow, such as Lucas-Kanade derivative of Gaussian, Lucas-Kanade, Farneback, and Horn-Schunck, are also employed. The Horn-Schunck~\citep{horn1981determining} method is selected since it provides magnitude and orientation features to create a binary magnitude-based mask localize and track individuals. The means, variances, and standard deviations are computed using these features to classify the individuals' abnormal behaviors.

In addition to using different pre-trained CNN models and optical flow estimators, different classifiers are examined, such as linear classifier, decision tree, and RF with cross-validation. The RF classifier is selected since it achieves better results in compared to the other classifiers. 

\textbf{Large-scale crowds:} The ResNet-50 and SqueezeNet pre-trained CNN models are used as a base network of the YOLOv2 object detection technique. We initialize the weights on ImageNet~\citep{krizhevsky2012imagenet}. Then, we fine-tune the model with stochastic gradient descent (SGD)~\citep{bottou1991stochastic} optimizer for 20 epochs with the learning rate of 0.001 and a minibatch of 8 frames. 

Similar to the small-scale crowd experiment, the RF classifier is also accountable for the abnormal behavior recognition in the temporal domain. Unlike in the small-scale crowd experiment, we train the RF classifier using only statistical features of the detected individuals with abnormal behaviors.

\subsection{Effectiveness Evaluation}
To evaluate the proposed methods, we evaluate them in both spatial and temporal domains. In the spatial domain, we consider the accuracy, precision, recall, F1 score, and Area Under the Curve (AUC) metrics as performance measures. The accuracy, precision, and recall metrics are defined in terms of True Positive (TP), True Negative (TN), False Positive (FP), and False Negative (FN) as the following:

\begin{equation} \label{eq1}
\begin{split}
Accuracy = \frac{TP+TN}{TP+TN+FP+FN}\\
Precision = \frac{TP}{TP+FP}\\
Recall = \frac{TP}{TP+FN}\\
F1 score = \frac{2 \times Precision \times Recall}{Precision + Recall}
\end{split}
\end{equation}

Receiver Operating Characteristics (ROC) curve~\citep{powers2020evaluation} is a plot of True Positive Rate (TPR) and False Positive Rate (FPR). The ROC curve represents the change of TPR and FPR over different thresholds. Thus, it is a powerful metric to evaluate a classifier. However, it is difficult to compare different classifiers using the ROC curve. Therefore, the AUC is used to compute the area under the ROC curve and compare the performance of the classifiers. The AUC scores range from zero to one. Stronger classifiers have higher AUC scores. \\

\begin{figure}[h]
  \includegraphics[scale=0.55]{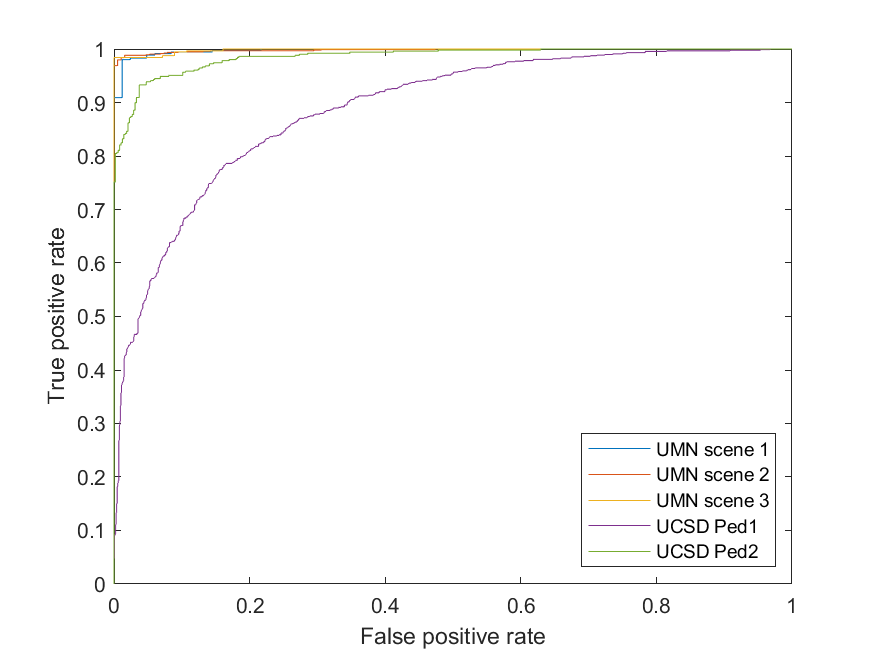}
  \caption{The ROC curves for the ResNet50 classifier on small-scale datasets.}
  \label{fig:small-scale-roc}
\end{figure}

\begin{figure}[h]
  \includegraphics[scale=0.55]{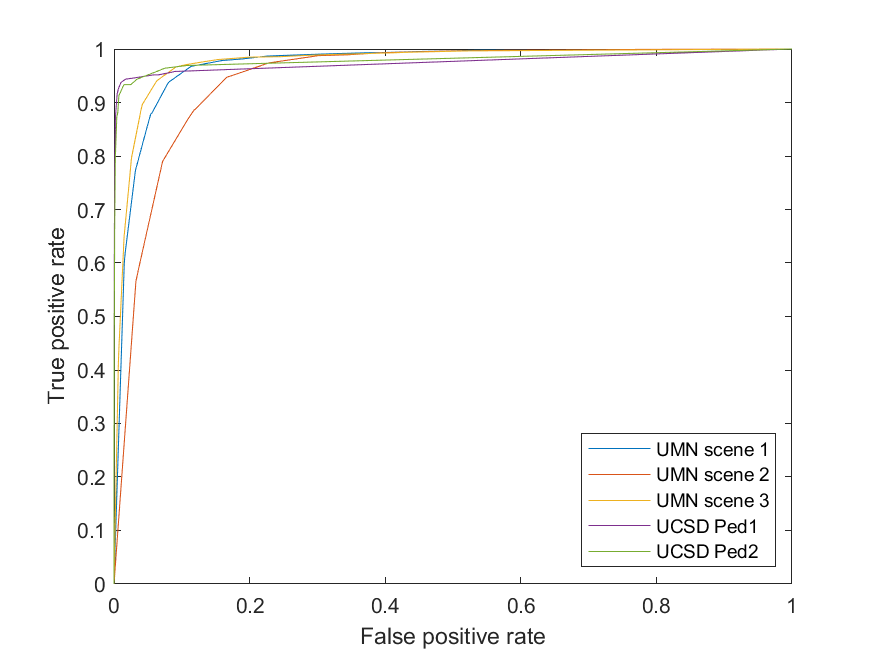}
  \caption{The ROC curves for the RF classifier on small-scale datasets.}
  \label{fig:RF-small-roc}
\end{figure}

\textbf{Small-scale crowds:} Table \ref{tab:small-ResNet} shows the frame detection quantitative results in the spatial domain using the small-scale crowd datasets. The ResNet-50 classifier achieves 99.77\% and 93.71\% of average AUCs among the scenes on UMN and UCSD datasets, respectively. Table \ref{tab:small-RF} shows the quantitative results in the temporal domain using the RF classifier on small-scale crowd datasets. Figures~\ref{fig:small-scale-roc} and \ref{fig:RF-small-roc} illustrate the ROC curves of our experiments using the ResNet-50 and the RF classifiers on UMN and UCSD datasets. In Figure~\ref{fig:qualitative-small}, samples of our qualitative results using the UMN and UCSD datasets are shown. One can notice that the proposed method detects and recognize the anomalies correctly in the datasets's testing samples.

To better illustrate the comparison with existing methods in~\citep{mehran2009abnormal},~\citep{cong2011sparse} and,~\citep{alafif2021generative}, Table~\ref{tab:UMN} shows that the proposed method yields better results using the UMN dataset. 

\begin{table}[h]
\centering
\caption{\label{tab:UCSD}The evaluation results on UCSD dataset. The percentages are AUCs.}
\resizebox{\columnwidth}{!}{%
\begin{tabular}{lcc}
\hline
Method & UCSD Ped1 & UCSD Ped2 \\\hline
MPPCA~\citep{kim2009observe} & 59.0\% & 69.3\% \\
Social Force[SF]~\citep{mehran2009abnormal} & 67.5\% & 55.6\% \\
SF+MPPCA~\citep{mahadevan2010anomaly} &68.8\% & 61.3\% \\
MDT~\citep{mahadevan2010anomaly} & 81.8\% & 82.9\% \\
Conv-AE~\citep{hasan2016learning} & 75.0\% & 85.0\% \\
Stacked RNN~\citep{luo2017revisit} & N/A & 92.2\% \\
Unmasking \citep{tudor2017unmasking} & 68.4\% & 82.2\%\\
Alafif et al.~\citep{alafif2021generative} & 82.81\% & 95.7\%\\
\textbf{Ours} & \textbf{88.87\%} & \textbf{98.55\%} \\
\hline
\end{tabular}}
\end{table}
\small
\begin{table}[h]
\centering
 \caption{\label{tab:UMN}The evaluation results on UMN dataset. Percentages are AUCs.}
 \resizebox{\columnwidth}{!}{%
 \begin{tabular}{lcl}
  \hline
  Method & UMN (\%) \\
  \hline
Optical-flow~\citep{mehran2009abnormal} & 84.0 \\
SFM~\citep{mehran2009abnormal})  & 96.0  \\
Sparse Reconstruction~\citep{cong2011sparse} & 97.0 \\
Alafif et al.~\citep{alafif2021generative} & 98.1 \\
\textbf{Ours} &  \textbf{99.8} \\
 \hline
 \end{tabular}
 }
\end{table}

\begin{table}[h]
\centering
\caption{\label{tab:small-RF}Our results of the RF classifier on small-scale crowd datasets.}
\resizebox{\columnwidth}{!}{%
\begin{tabular}{lccccc}
\hline
Dataset     & Accuracy (\%) & Precision (\%) & Recall (\%) & F1 (\%)    & AUC (\%)\\\hline
UMN scene 1 &   89.69       &   99.33        &   88.36     &   93.52    & 97.11   \\
UMN scene 2 &   81.51       &   99.04        &   76.81     &   86.51    & 94.66   \\
UMN scene 3 &   93.69       &   99.47        &   93.66     &   96.47    & 97.62   \\
UCSD Ped1   &   99.5        &   99.6         &   99.89     &   99.75    & 97.56   \\ 
UCSD Ped2   &   99.6        &   99.82        &   99.78     &   99.79    & 97.99   \\ 
\hline
\end{tabular}}
\end{table}

Table~\ref{tab:UCSD} reports a performance comparison of the proposed method with the existing methods~\citep{kim2009observe, mehran2009abnormal, mahadevan2010anomaly, hasan2016learning, tudor2017unmasking, alafif2021generative} using the UCSD dataset. It is clearly shown that the proposed method achieves higher AUCs using USCD Ped1 and Ped2 scenes.\\ 

\begin{table*}[ht]
\centering
\caption{\label{tab:Hajj-yolo-results}The experimental results of track assignment and IOU detections using YOLOv2 on HAJJv2 dataset.}
\resizebox{\textwidth}{!}{%
\begin{tabular}{lcccccccc}
\hline
& \multicolumn{4}{c}{Track Assignment} & \multicolumn{4}{c}{IOU} \\
\cline{2-9}\\
Video No.                & Accuracy (\%) & Precision (\%) & Recall (\%) & F1 (\%) & Accuracy (\%) & Precision (\%) & Recall (\%) & F1 (\%) \\ \hline
10 (Arafat)              & 89.86         & 96.41          & 54.27       & 69.44   & 91.25         & 56.16          & 31.61       & 40.45   \\
12 (Tawaf)               & 96.66         & 98.84          & 2.92        & 5.67    & 96.87         & 4.65           & 0.14        & 0.27    \\
9 and 11 (Jamarat)       & 89.26         & 96.44          & 3.85        & 7.40    & 89.53         & 14.24          & 0.57        & 1.09    \\
2, 3, 5, 7, and 8 (Masaa)& 91.30         & 78.19          & 50.93       & 61.69   & 93.23         & 51.66          & 33.65       & 40.75   \\ \hline
Average                  & 91.77         & 92.47          & 27.99       & 36.05   & 92.72         & 31.68          & 16.49       & 20.62    \\ \hline
\end{tabular}}
\end{table*}

\begin{table*}[ht]
\centering
\caption{\label{tab:Hajj-RF-results}Abnormal behaviors recognition results using the RF classifier on HAJJv2 dataset.}
\resizebox{\textwidth}{!}{%
\begin{tabular}{lccccc}
\hline
Video No. & Accuracy (\%) & Precision (\%) & Recall (\%) & F1 (\%) & AUC (\%) \\\hline
10 (Arafat)      & 63.65         & 64.81          & 58.30       & 61.38   & 91.18    \\ 
12 (Tawaf)      & 62.91         & 64.59          & 52.59       & 57.80   & 75.11    \\ 
9 and 11 (Jamarat)     & 96.86         & 33.92          & 33.03       & 33.47   & 57.92    \\ 
2, 3, 5, 7, and 8 (Masaa)   & 56.44         & 37.97          & 32.81       & 35.20   & 80.12    \\\hline
Average   & 75.42         & 50.32          & 44.15       & 46.96   & 76.08    \\\hline
\end{tabular}}
\end{table*}

\begin{table}[ht]
\footnotesize
\centering
\caption{\label{tab:Hajjv1-vs-Hajjv2}A comparison table of abnormal behaviors detection performance with the recent existing methods using HAJJv1 dataset against existing methods.}
\resizebox{\columnwidth}{!}{%
\begin{tabular}{lccccc}
\hline
Model                                       & Accuracy (\%) & Precision (\%) & Recall (\%) & F1(\%) & AUC (\%)  \\\hline
\cite{alafif2021generative}  & 65.10         & 61.48          & 80.30       & N/A     & 79.63\\ 
YOLOv2 (Ours)                                        & \textbf{95.67}         & 9.421          & 28.82       & 10.99 & N/A    \\%
 RF (Ours)                                         & 34.48         & 9.910          & 11.22       & 10.39 & 66.56
\\
\hline
\end{tabular}}
\end{table}

\begin{figure}[h]
  \includegraphics[scale=0.55]{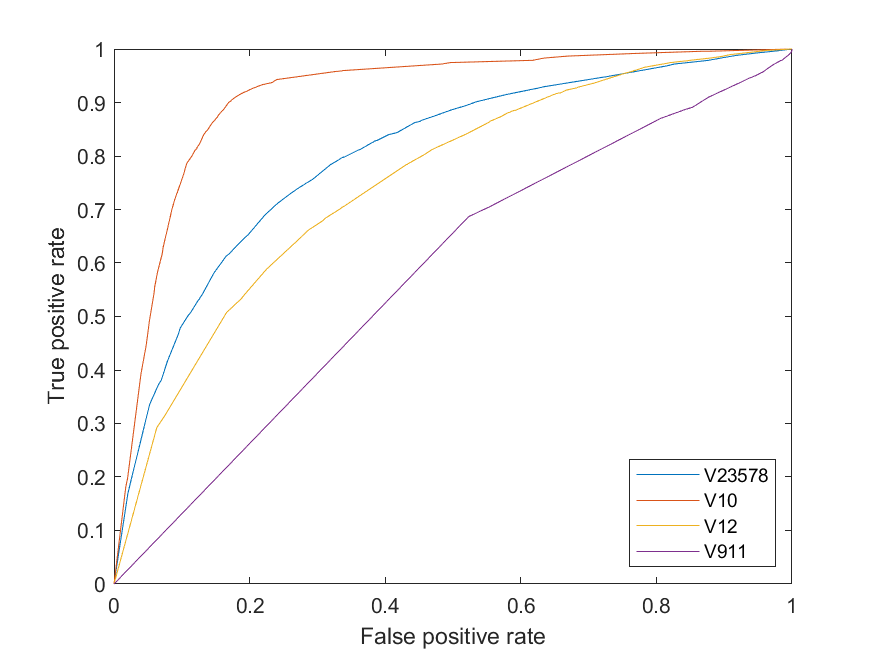}
  \caption{The ROC curves for the RF classifier on HAJJv2 dataset.}
  \label{fig:RF-Hajj-roc}
\end{figure}

\textbf{Large-scale crowds:} 
We evaluate our method in the spatial and temporal domains on large-scale crowds using HAJJv1 and HAJJv2 datasets. The spatial domain is evaluated using two criteria, track assignment and Intersection Over Union (IOU). The results of track assignment are computed using Kalman filter assignment results for each detected object. Nevertheless, it is not important if the detected pixels match most of the labeled pixels exactly. Thus, we use the IOU to evaluate the YOLOv2 detector. The IOU is a powerful evaluation metric to evaluate the detection of objects as it is commonly used in the computer vision community. It finds the overlap ratio between the ground-truth and detected boxes. Then, using a 50\% threshold of the overlapping boxes, we compute the TP, FP, and FN. The accuracy of the YOLOv2 is computed on the pixels level. A pixel is considered TN if no TP, FP, and FN pixel is detected by the detector at this pixel. It is observable that the accuracy cannot report the performance well. Since YOLOv2 doesn't detect any anomalies at most of the frames' pixels, and since the majority of frames' pixels don't contain abnormal behaviors, the TN number is increased. That affects the accuracy calculation and neglects the values of TP, FP, and FN. \\
Table~\ref{tab:Hajj-yolo-results} and \ref{tab:Hajj-RF-results} show our quantitative results in the spatial and temporal domains using HAJJv2 dataset. Our fine-tuned pre-trained ResNet-50 model with the YOLOv2 detector achieves 92.47\% precision using track assignment and 31.68\% using IOU. In the temporal domain, the RF classifier achieves 76.08\% of AUC for abnormal behaviors recognition on the HAJJv2 dataset. Figure~\ref{fig:qualitative-Hajj} shows the qualitative results for the proposed method on the HAJJv2 dataset. Figure~\ref{fig:RF-Hajj-roc} demonstrates the ROC curves for the RF classifier. A quantitative comparison with the work in~\citep{alafif2021generative} using the HAJJv1 dataset is also provided in Table~\ref{tab:Hajjv1-vs-Hajjv2}.

\begin{table}[h]
\centering
\caption{\label{tab:small-ResNet}Our results using the ResNet-50 classifier on public and benchmark small-scale crowd datasets.}
\resizebox{\columnwidth}{!}{%
\begin{tabular}{lccccc}
\hline
Dataset      & Accuracy (\%) & Precision (\%) & Recall (\%) & F1 (\%)   & AUC (\%)\\\hline
UMN scene 1  &   97.93       &   98.31        &   99.15     & 98.73     &  99.73   \\
UMN scene 2  &   98.49       &   99.36        &   98.82     & 99.09     &  99.79   \\
UMN scene 3  &   98.07       &   99.46        &   98.41     & 98.93     &  99.77   \\
UCSD Ped1    &   75.72       &   64.72        &   89.31     & 75.05     &  88.87   \\  
UCSD Ped2    &   94.14       &   96.29        &   92.11     & 94.15     &  98.55   \\  
\hline
\end{tabular}}
\end{table}

\section{Discussion}\label{sec:Dis}
 The proposed methods are robust in detecting and recognizing individuals with abnormal behaviors in small-scale and large-scale crowd videos. The results show that the small-scale crowd method achieves a great performance in comparison with the state-of-the-art techniques. Although the small-scale method outperforms other existing techniques, it shows an unsatisfactory performance when using the UCSD Ped1 dataset. Several factors contributed to this, including the low resolution of the frame, the camera viewing, the shadows cast by trees, and the low illumination. In the large-scale crowds, we still haven't achieved an excellent performance, yet using the HAJJv2 dataset since the videos in the dataset are very challenging. The challenges are represented by far camera viewing and heavy partial and full occlusions with a huge number of individuals. Figures~\ref{fig:HAJJ-Results-v12} and \ref{fig:HAJJ-Results-v9} show some of the challenges in Tawaf and Jamarat scenes which are considered the hardest scenes for the classifiers to classify the individuals with abnormal behaviors. During annotating and labeling the abnormal behaviors using these scenes, much more human attention and focus were required since they contain a massive number of individuals moving in one spot with heavy partial occlusions and far camera views. On the other hand, the easiest scenes for the annotators, labelers, and classifiers to classify are in the Masaa scenes in which most of the individuals are classified correctly since these videos are captured from a closed camera view and have a moderate number of partial occlusions. Therefore, these factors definitely contribute to the performance of the abnormal behavior detection and recognition classifiers. Much more future work is required 
 to better detect and recognize individuals with abnormal behaviors in large-scale and massive crowds.

\section{Conclusion}\label{sec:Con}
In this research work, we first introduce the annotated and labeled large-scale crowd abnormal behaviors Hajj dataset (HAJJv2). Second, we propose two methods of hybrid CNNs and RFs to detect and recognize Spatio-temporal abnormal behaviors in small-scale and large-scale crowd videos. In small-scale crowd videos, a ResNet-50 pre-trained CNN model is fine-tuned to check every frame, whether it is normal or abnormal in the spatial domain. If abnormal behaviors are found, a motion-based individual detection using magnitude and orientation features of Horn-Schunck optical flow is employed to create a binary magnitude-based mask to localize and track individuals with abnormal behaviors. In large-scale crowd videos, a Kalman filter is employed to predict and track the detected individuals in the next frames. Then, means, variances, and standard deviations statistical features are computed and fed to the RF to classify individuals with abnormal behaviors in the temporal domain. In the large-scale crowd videos, we fine-tune the ResNet-50 model using the YOLOv2 object detection technique to detect individuals with abnormal behaviors in the spatial domain. Our method using public benchmark small-scale crowd datasets achieves 99.77\% and 93.71\% of AUCs respectively on the UMN and UCSD datasets, while the method in large-scale crowd achieves 76.08\% of average AUC using the HAJJv2 dataset. Our methods outperform state-of-the-art methods using the small-scale crowd datasets with a margin of 1.67\%, 6.06\%, and 2.85\% on the UMN, UCSD Ped1, and UCSD Ped2 datasets, respectively. It also achieves a satisfactory result in large-scale crowds. Still, lots of work is needed to increase the effectiveness of abnormal behavior detection and recognition in large-scale crowded scenes due to their challenges. Most of the current research works only uses small-scale crowded scenes in which the abnormal behaviors can be extracted and classified easily. In the future, our work will be focused on large-scale crowds. We will incorporate an attention mechanism and fusion strategies to enhance the performance. This work can potentially help researchers to study and apply it in different contexts of crowd scenes such as in airports, stadiums, and marathons. It can be also extended in the manufacturing industries ~\citep{patel2018erel} by inspecting and detecting unusual behaviors of defective manufactured goods and products on a production line. Examples and features of the products' unusual behaviors are required to be collected, extracted, and learned by a classifier to achieve high performance.

\section{Acknowledgement}
The authors extend their appreciation to the Deputyship for Research and Innovation, Ministry of Education in Saudi Arabia for funding this research work through the project number (227).

\bibliography{refNew.bib}
\bibliographystyle{cas-model2-names}
\end{document}